\documentclass[preprint]{elsarticle}
\pdfoutput=1
\usepackage{lineno,hyperref}
\modulolinenumbers[5]

\journal{Medical Image Analysis}

\biboptions{authoryear}

\graphicspath{{images/}}

\usepackage{amsmath}
\usepackage{amsfonts}
\usepackage{breqn}

\usepackage{nicefrac}
\usepackage{kbordermatrix}

\usepackage{algorithmic}

\usepackage[english]{babel}
\usepackage{bm}

\usepackage{longtable}
\usepackage{rotating}
\usepackage{array}
\usepackage{multicol}
\usepackage{multirow}

\usepackage{algorithm}

\newcommand{\matr}[1]{\mathbf{#1}}

\usepackage{float}
\usepackage{subfigure}

\restylefloat{table}

\usepackage{csvsimple}
\usepackage{siunitx}
\usepackage{tikz}

\usepackage{soul}

\begin{document}

\begin{frontmatter}

\title{Hubless keypoint-based 3D deformable groupwise registration}



\author[creatis]{R. Agier}
\author[creatis]{S. Valette\corref{mycorrespondingauthor}}
\ead{sebastien.valette@creatis.insa-lyon.fr}
\cortext[mycorrespondingauthor]{Corresponding author}
\author[creatis]{R. K\'{e}chichian}
\author[creatis,hcl]{L. Fanton}
\author[creatis]{R. Prost}

\address[creatis]{Université de Lyon, INSA Lyon, Université Claude Bernard Lyon 1, UJM-Saint Etienne CNRS, Inserm, CREATIS UMR 5220, U1206, F69621 LYON, France}
\address[hcl]{Hospices Civils de Lyon, GHC, H\^{o}pital Edouard-Herriot, Service de m\'{e}decine l\'{e}gale, 69003 LYON, FRANCE.}

\begin{abstract}
	We present a novel algorithm for Fast Registration Of image Groups (FROG), applied to large 3D image groups. Our approach extracts 3D SURF keypoints from images, computes matched pairs of keypoints and registers the group by minimizing pair distances in a hubless way i.e. without computing any central mean image. Using keypoints significantly reduces the problem complexity compared to voxel-based approaches, and enables us to provide an in-core global optimization, similar to the Bundle Adjustment for 3D reconstruction. As we aim to register images of different patients, the matching step yields many outliers. Then we propose a new EM-weighting algorithm which efficiently discards outliers. Global optimization is carried out with a fast gradient descent algorithm. This allows our approach to robustly register large datasets. The result is a set of diffeomorphic half transforms which link the volumes together and can be subsequently exploited for computational anatomy and landmark detection. We show experimental results on whole-body CT scans, with groups of up to 103 volumes. On a benchmark based on anatomical landmarks, our algorithm compares favorably with the star-groupwise voxel-based ANTs and NiftyReg approaches while being much faster. We also discuss the limitations of our approach for lower resolution images such as brain MRI.
\end{abstract}


\end{frontmatter}


\section{Introduction}
\label{sec:intro}

Registration of several images together, also known as groupwise registration, is nowadays most often carried out for human brain studies \citep{jenkinson2002improved}. Whole body studies remain rare as they still raise significant problems \citep{XU2016BENCHMARK}. Advances in imaging techniques are constantly increasing the number and size of 3D images in hospital databases, hence the need for low complexity groupwise image registration techniques. Instead of dense voxel based registration, sparse keypoint matching, extracted from the images is a promising approach. Keyoints are extracted with their location and description vector. Points from different images are paired by comparing their description vectors. This is a great challenge when shifting from intrapatient registration to interpatient registration, as the human anatomy exhibits a large variability.

In this paper we propose a novel groupwise registration approach aimed at large image databases which is able to register high resolution images, such as whole body CT scans shown in Figure \ref{fig:wb}. As human anatomy exhibits high variability (e.g. men vs. women), our algorithm is hubless, i.e. it does not use any central reference during registration. Inspired by advances in the computer vision field, our algorithm exploits keypoint detection and matching, which bring speed and robustness. Some paired points are erroneous (outliers) and we devised an algorithm for robustness against these outliers, even when they are very frequent. Potential applications range from computational anatomy to forensic anthropology and robust patient-specific model construction. An example of forensic application is probabilistic sex diagnosis using worldwide variability in hip-bone measurements \citep{DSP}. Moreover, the ever-increasing amount of medical images stored in Picture Archiving and Communication Systems (PACS) in hospitals offers a great opportunity for big data analysis. Screening huge image groups could improve diagnosis and could be of critical help for computational anatomy. Recent advances in machine learning \citep{michalski2013machine} or atlas based-approaches \citep{Iglesias2015survey} have pushed for efficient big data analysis tools. Yet the modest size of current annotated medical image databases limits the use of these approaches for medical imaging. As a consequence, whole-body groupwise registration could bridge the gap between big data and organ localization, segmentation, and computational anatomy. One known limitation for our approach is that the number of extracted keypoints should be as high as possible to obtain the best accuracy.

The paper is organized as follows: section \ref{sec:rel} presents previous works related to registration, keypoints and Bundle Adjustment. Section \ref{sec:chacont} summarizes the key contributions of our hubless approach. Section \ref{sec:method} explains our approach in detail while section \ref{sec:results} shows results obtained with our approach, as well as comparisons with both NiftyReg and ANTs star-groupwise algorithms. We conclude the paper in section \ref{sec:discussion}.

\section{Related works}
\label{sec:rel}
\subsection{Image registration}
\label{sec:registration}

Image registration, also known as image matching, fusion or warping, consists of finding a transform $\tau$ between two or more images, mapping any point $p$ from a source image to a position $\tau(p)$ in the target image. It is a crucial step in many medical applications, such as longitudinal studies \citep{scahill2003longitudinal}, radiotherapy planning \citep{keall2005four}, brain studies \citep{jenkinson2002improved}, atlas-based segmentation \citep{lotjonen2010fast}, image reconstruction \citep{huang2008three} and microscopy \citep{vercauteren2007non}. We present here a brief state of the art, and we refer the reader to \cite{Brown:1992:SIR:146370.146374, maintz1998survey, sotiras2013deformable} for thorough reviews of existing approaches.

\subsubsection{Pairwise \& Groupwise registration}

Pairwise registration consists of matching one moving image to a fixed image. Groupwise registration consists of registering a whole set of images together.

\begin{figure}

\subfigure[Star]{\label{fig:star_groupwise} \includegraphics[width=.36\linewidth]{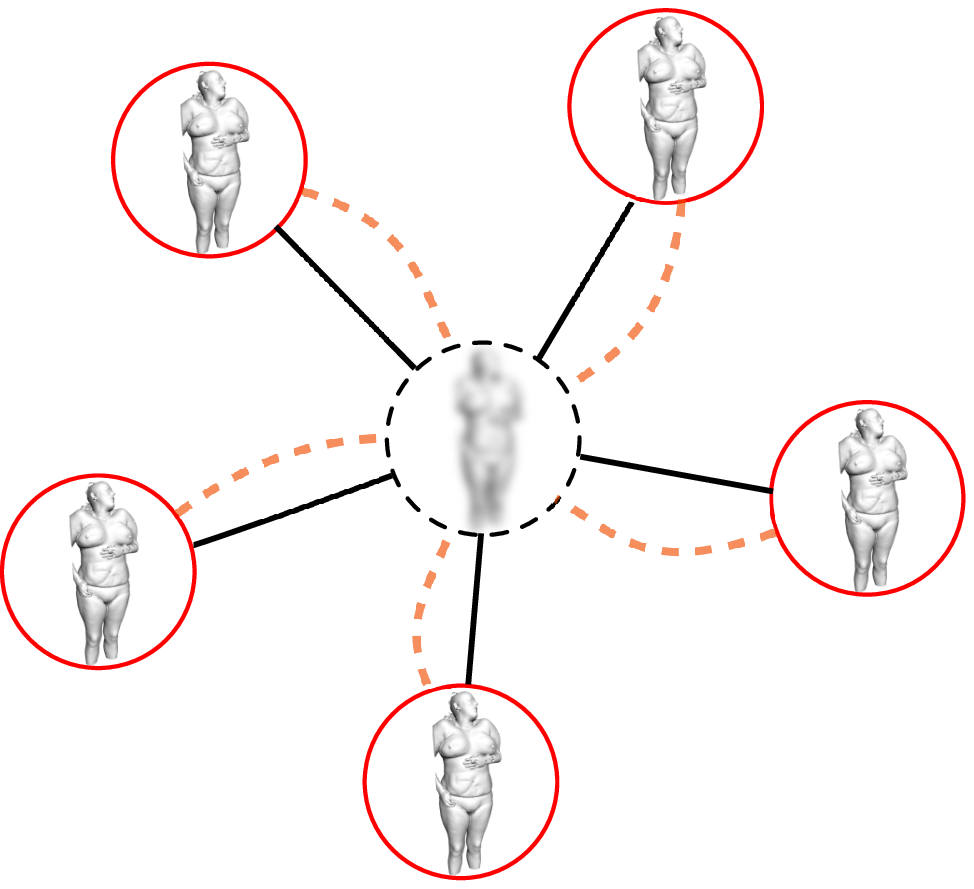}}
\subfigure[Tree]{\label{fig:tree_groupwise}
	\includegraphics[width=.28\linewidth]{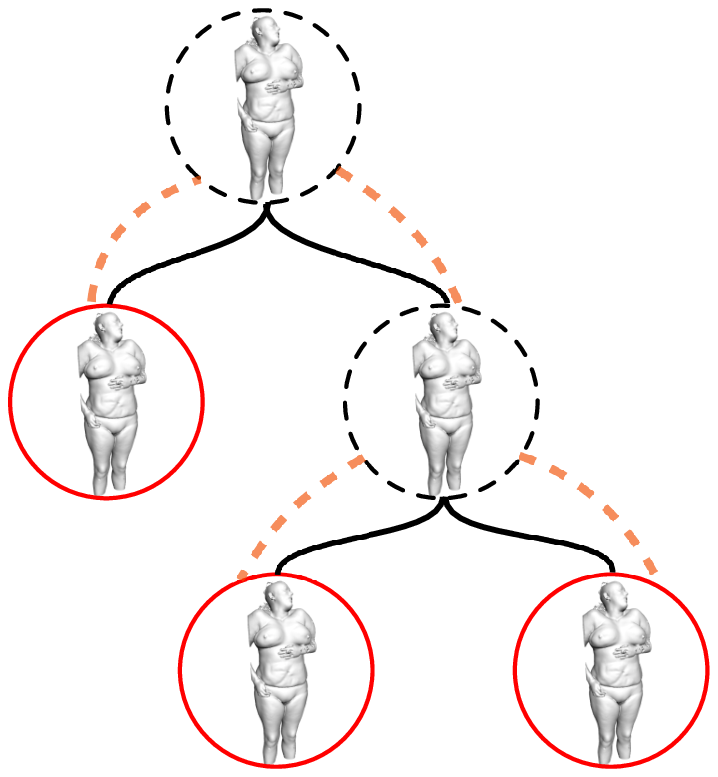}}
\subfigure[Hubless] {\label{fig:hubless_groupwise}\includegraphics[width =.31\linewidth]{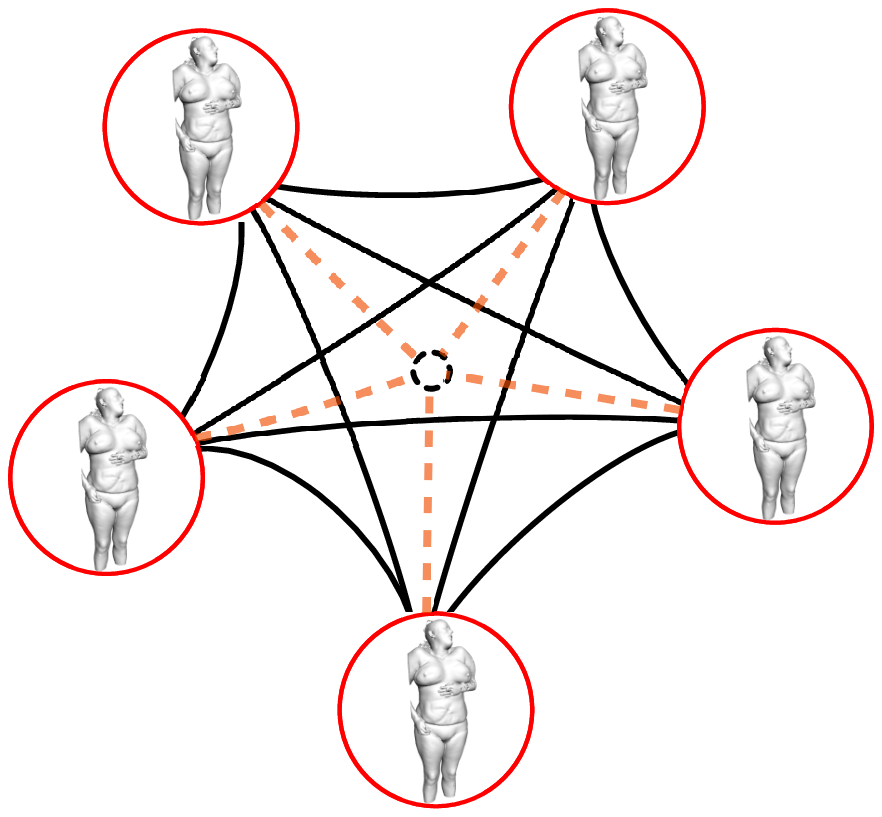}}

	\caption{ Different groupwise registration methods. Images are outlined in red, reference images are outlined in dotted black, transforms in dashed orange and observations in black. For dense registrations, observations are generally voxelwise difference, cross-correlation or mutual information. For sparse registrations, observations are usually keypoint match distances. Transforms are optimized via the minimization of an energy function driven by the observations. (a) star groupwise registration (1 reference, $n$ transforms and $n$ observations). (b) tree-groupwise registration (multiple references, $n-1$ transforms, $n-1$ observations). (c) hubless registration, with no reference image but one abstract common space, $n$ half-transforms and at most $\frac{n(n-1)}{2}$ observations.}
	\label{fig:hubless}
\end{figure}

One can categorize groupwise registration approaches using graph theory, as shown in Figure \ref{fig:hubless}, where vertices represent images and edges represent transforms. Then, transforming any image to any other image requires at least a graph spanning all vertices. Then at least $n-1$ transforms are needed for $n$ images.

\begin{itemize}
	\item Star groupwise methods register each image of the set against a reference image. In this case the graph is a star graph (Figure \ref{fig:star_groupwise}). The output is a set of half transforms $\tau_i$, mapping each image $i$ to the common space. After registration, the transform $\tau_{i \rightarrow j}$ from one image $i$ to another image $j$ is a composition of the two half transforms $\tau_i$ and $\tau_j$: $\tau_{i \rightarrow j} = \tau_j^{-1} \circ \tau_{i\vphantom{j}}^{\vphantom{-1}}$. These methods are biased by the choice of the initial reference or template image. This bias takes two forms : the intensity bias and the shape bias. \cite{guimond00} reduced these biases by iteratively updating the reference image, taking into account the average of each transformed image, and the average of image transformations. Later, \cite{joshi2004unbiased} improved the theoretical foundations of this approach via diffeomorphism. An example of application for brain pediatric studies is shown in \cite{fonov11}. The publicly available softwares ANTs\citep{Avants2008ants}, Elastix\citep{klein10} and NiftyReg\citep{modat2008} belong to the star groupwise category.
	
	\item Tree groupwise methods use a spanning tree graph and multiple references \citep{wu2011sharpmean}, as shown by Figure \ref{fig:tree_groupwise}. Variability can then be distributed across several references instead of only one. 
	
	\item Hubless methods use an abstract common space, but do not use any reference image: while star and spanning tree graphs are minimal graphs, hubless methods use a dense graph to benefit from more constraints, as shown by Figure \ref{fig:hubless_groupwise}. When registering $n$ images, up to $n(n-1)/2$ observations can be used. Observations can be pairwise registrations, or local matches. \cite{hamm2010} proposed GRAM, a hubless registration method, based on manifold learning. This algorithm requires dense pairwise registration of all image pairs, which is very time consuming. As an example, the authors report a processing time of 24 hours on a cluster to register 416 low resolution images ($ 68 \times 56 \times 72 $). \cite{ying2014hierarchical} proposed  a similar approach, using geodesic graph shrinkage. A recent work on groupwise nonrigid registration was also proposed by \cite{wu2012feature}, but is dedicated to the processing of brains acquired with MRI, as it requires a segmentation of the image into White Matter, Grey Matter and Ventricular Cerebrospinal Fluid. More recently, the approach proposed in \cite{agier2016hubless} uses no reference at all. This allows handling partially matching body parts and high variability but is restricted to rigid transforms. 
\end{itemize}

Groupwise registration has most often been used for brain analysis, as the brain shape exhibits low variations between subjects overall. Whole body studies such as carried out by \cite{Suh2011} remain rare. The important advantage of hubless approaches becomes clear when averaging a group of images is a problem. A good example is to register full body images, where computing the average between men and women is inconsistent in several regions.

\subsubsection{Rigid \& Deformable Registration}

Rigid registration \citep{ashburner2003rigid} consists of computing a linear transform between two images. The transform can be a translation or a more expressive one such as a combination of translation, rotation, or scaling. Rigid transforms can be used for Procrustes Analysis \citep{schonemann1966generalized}, Generalized Procrustes Analysis \citep{Bartoli2013}, mosaic assembly or as an initialization step for nonrigid transform.
Nonrigid (deformable) registration consists of finding a free-form transform able to express local variability, which cannot be done with a rigid transform. Nevertheless, the transform cannot be arbitrary, and should generally be smooth and invertible.

\subsubsection{Dense \& Sparse Registration}

Registration approaches can also be split in two classes, depending on how they process the input data:

Dense registration estimates displacement on the entire image domain \citep{Avants2008ants, modat2008, modat2014}. The most common approaches use intensity difference between images but there are many other criteria such as cross-correlation\citep{lewis1995fast} or mutual-information\citep{pluim2003mutual}. The output is a displacement vector for each input voxel. The key difficulties with dense registration are high computational cost and an ill-posed optimization problem, which generally needs explicit regularization\citep{robinson2004fundamental}. As an example, the Advanced Normalization Tools (ANTs) framework \citep{Avants2008ants} registers two thoracic images in about 2 hours on our test machine.

Sparse registration, instead of working on a dense grid, uses only point sets.
Compared to dense registration, sparse registration is much faster\citep{allaire2008full,cheung2009n} but may yield less accurate results  because points do not always span the whole space. For nonrigid registration, sparse transforms need an interpolant, such as splines \citep{szeliski1997spline} or radial basis functions \citep{fornefett2001radial}. The interpolant inherently provides a smooth solution which helps regularizing the solution. Point sets can be of different types:
\begin{itemize}
	\item Reference landmarks which are distinguishable anatomic structures manually placed on the image by an expert. This task is usually carried out by physicians and can be time consuming. These are used in \citep{wang06,li12}
	\item Reference landmarks automatically placed on the image by an algorithm, such as proposed by Zhu et al.\citep{zhu13}
	\item Vertices of a surface mesh representing the object to register, such as proposed in \citep{rasoulian12}
	\item Keypoints, automatically detected and placed in the image. One important difference between reference landmarks and keypoints is that keypoints are not defined by anatomical definitions but by mathematical properties of the image. Hence, although it is interesting to find the same keypoints in each image, this is never the case in practice, and the number of extracted points from each image is not a priori known. Next section (\ref{sec:keypoints}) gives more details on keypoints.
\end{itemize}


Some works use a combination of both dense and sparse approaches. As an example, \cite{wu2012feature} uses driving voxels to speed up a dense method.

\subsection{Keyoints}
\label{sec:keypoints}
During the last decades keypoints \citep{harris1988combined} have successfully been used for object recognition \citep{lowe1999object}, action recognition \citep{wang2011action}, robotic navigation with Simultaneous Localization and Mapping \citep{karlsson2005vslam} and panorama creation \citep{anguelov2010google}. They aim at being fast while reducing the amount of data to process, mainly to deal with real time processing or tasks involving large amounts of data.
Their suitability to medical imaging has been evaluated in \citep{lopez1999evaluation}, and various applications have been proposed in this context such as image annotation \citep{datta2005content}, retrieval \citep{zheng2008interest} and segmentation \citep{KECH-14,wachinger18}. Initially designed for 2D images, keypoint approaches have been extended to process 3D medical images by \citep{allaire2008full, agier2016hubless, cheung2009n, SIFT3D}.

The keypoint extraction pipeline is the following: first, for each image, a detector extracts locations exhibiting important features such as corners (first derivative analysis \citep{harris1988combined}) or blobs (second derivative), most often via fast approximations of theses derivatives. As an example, Scale Invariant Feature Transform (SIFT) uses difference of Gaussians \citep{lowe2004distinctive} and Speeded Up Robust Features (SURF) use integral images \citep{bay2006surf}.
Afterwards, the neighborhood of each location is summarized into a compact feature vector, called descriptor. SIFT populates the feature vector with orientation histograms, while SURF stores local texture information. 

Efficient registration can be performed as follows: (1) keypoints are extracted from the input images, with locations and descriptions. (2) Keypoints from different images are paired by comparing their feature vectors. Note that the number of keypoints may vary between images. This results in a set $\mathcal{M}$ of paired points. (3) Registration is performed by minimizing the distance between all paired points of this set \citep{mikolajczyk2005performance}.

The use of paired points combined with sparse registration approaches leads to very fast registration, in opposition to dense registration. Moreover, as keypoints have to be extracted only once per image, repeated registration with the same reference image is very efficient.

Finally, keypoints are rarely used for groupwise registration, most of the previous approaches use anatomical reference landmarks to perform groupwise registration. Few papers deal with keypoints, such as Zhang et al. \citep{zhang12}. Note that \citep{zhang12} is restricted to 2D images.

\subsection{Outlier rejection}
\label{sec:outliers}
A drawback of using keypoints is the significant number of mismatches (outliers) in the paired point set $\mathcal{M}$. Outliers are pairs of keypoints with similar descriptors describing regions which are actually different and therefore should not be paired. Decreasing the image quality (using low end cameras and webcams used in consumer electronics devices) potentially increases the outlier ratio to more than 30\% of the set \citep{labe2004geometric} (37\% in \citep{mikolajczyk2002affine}).

In \cite{Rangarajan97}, a spatial mapping and the one-to-one correspondences (or homologies) between point features extracted from the images try to reject non-homologies as outliers. In \cite{Chui03}, an alternating optimization algorithm successively updates the points correspondence matrix and the transformation function while gradually reducing the temperature in a deterministic annealing. The points correspondence matrix handles outliers.

In \cite{Myronenko10}, a probabilistic method called Coherent Point Drift (CPD) fits the first point set centroids as a mixture of Gaussian with an additional uniform distribution to account for outliers point set centroids to the data points by likelihood maximization. The method forces the centroid to move coherently as a group to preserve the topological structure.

M-estimators \citep{huber2011robust} provide a good way to mitigate the influence of outliers in this context. During optimization, M-estimators provide an adaptive weighting function which adjusts the contribution of each pair by computing robust statistics on $\mathcal{M}$. Each point pair can then be classified as inlier or outlier (soft classification) and weighted accordingly. For this aim, one needs an estimation of the inlier contribution variance, computed via Median Absolute Deviation (MAD). Note that the maximum outlier ratio that MAD can process (the breakdown point \citep{donoho1983notion}), is 50\%. Hence the breakdown point of the M-estimator is also 50\%.\\

\subsection{Bundle Adjustment}

Bundle adjustment (BA) is a class of computer vision approaches which allows reconstruction of a unique 3D scene from multiple 2D views by back projection of keypoints in the 3D scene and global optimization \citep{triggs2000bundle}. It allows efficient tracking even with low-end cameras \citep{karlsson2005vslam}. Recent BA implementations can manage large amounts of data and successfully reconstruct large parts of a city \citep{frahm2010building}.

BA jointly optimizes point positions, camera positions and camera parameters. Because of (1) the non-linearity of projective geometry \citep{hartley2003multiple}, (2) mismatches and (3) the M-estimator, optimization is high-dimensional and non-convex \citep{NIPS2014_5486}. As a consequence, particular attention has to be paid to the optimization method. Algorithms such as Levenberg-Marquart optimization \citep{more1978levenberg} can be used to avoid converging to a local minimum. BA also needs a good initialization to remain close to the global minimum at any time.

\section{Challenges and contributions}
\label{sec:chacont}
\begin{figure}
	\begin{minipage}[c]{.6\linewidth}
		\includegraphics[width=\linewidth]{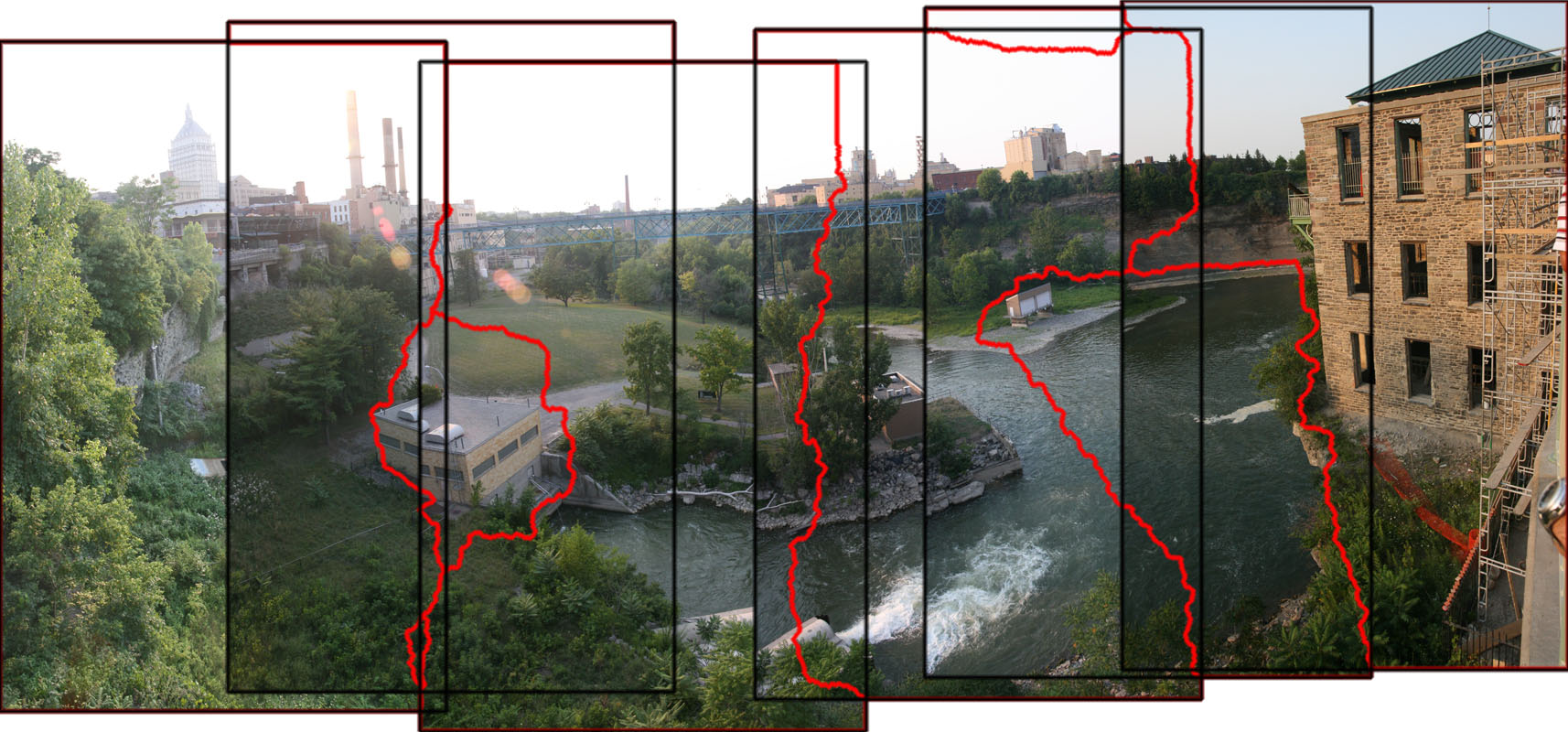}
	\end{minipage}
	\hfill
	\vspace{10mm}
	\begin{minipage}[c]{.38\linewidth}
		\tiny
		$\begin{bmatrix}
		\bullet & \bullet & \bullet& & & & \\
		\bullet & \bullet & \bullet & & & & \\
		\bullet & \bullet & \bullet & \bullet & & & \\
		&  & \bullet & \bullet & \bullet & & \\
		& &  & \bullet & \bullet & \bullet & \bullet\\
		&  & &  & \bullet & \bullet & \bullet \\
		& & & & \bullet & \bullet & \bullet  \\
		\end{bmatrix}
		$
	\end{minipage}
	\begin{minipage}[c]{.6\linewidth}
		\includegraphics[width=\linewidth]{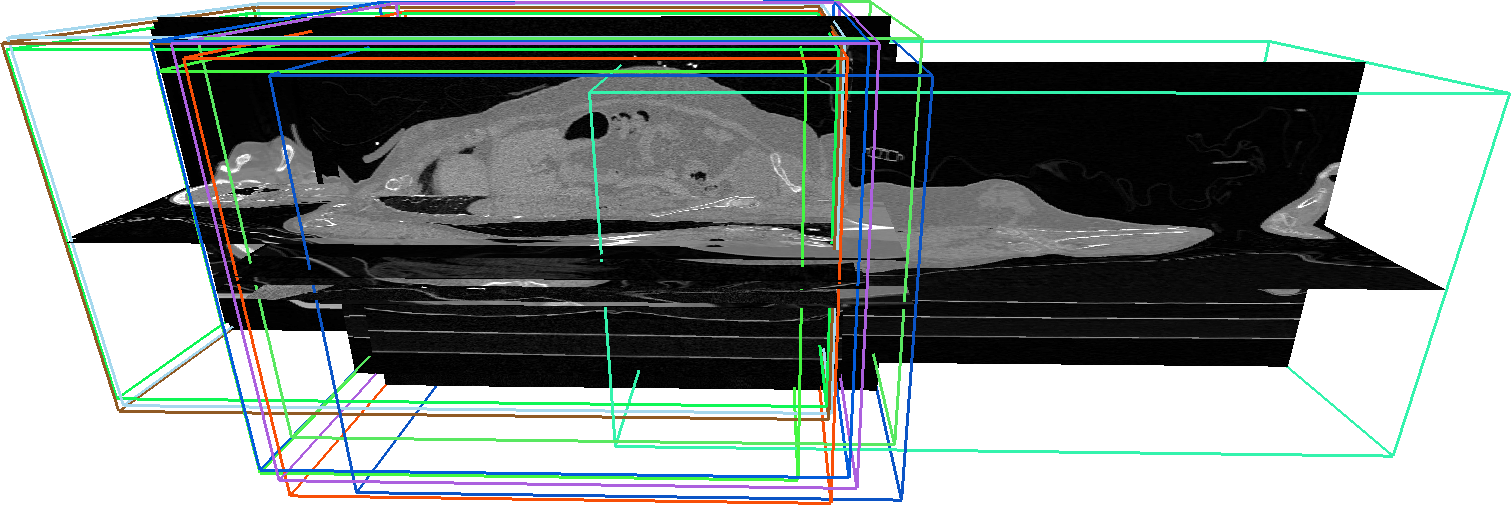}
	\end{minipage}
	\hfill
	\begin{minipage}[c]{.38\linewidth}
		\tiny
		$\begin{bmatrix}
		\bullet & \bullet & \bullet & \bullet & \bullet & \bullet & \bullet \\
		\bullet & \bullet & \bullet & \bullet & \bullet & \bullet & \bullet \\
		\bullet & \bullet & \bullet & \bullet & \bullet & \bullet & \bullet \\
		\bullet & \bullet & \bullet & \bullet & \bullet & \bullet & \bullet \\
		\bullet & \bullet & \bullet & \bullet & \bullet & \bullet & \bullet \\
		\bullet & \bullet & \bullet & \bullet & \bullet & \bullet & \bullet \\
		\bullet & \bullet & \bullet & \bullet & \bullet & \bullet & \bullet \\
		\end{bmatrix}
		$
	\end{minipage}
	
	\caption{Sparsity induced by a panoramic reconstruction with BA (top) compared to our application (bottom). Overlap matrices of each problem are given on the right. The top image is provided by Noso1, via Wikimedia Commons.}
	\label{fig:sparsity}
\end{figure}

Our paper provides a unified solution to register in a nonrigid way a whole image bundle at once. It can be seen as a complete graph registration because each image is linked with all other images (figure \ref{fig:hubless_groupwise}). Several challenges need to be addressed: 

\begin{itemize}
	
	\item \emph{Amount of data: } As we want to process many images in reasonable time, we cannot use dense registration: each image from our test database contains about $512^3$ voxels, each voxel being encoded with 32 bits.  Depending on the experiments, 20 to more than 100 images are used, representing more than 50GB of image data. Current approaches process a lower number of images, as in \citep{LPBA40}. On top of a hubless approach as in \citep{wu2012feature}, we need a compact solution which avoids reading voxel data and allows us to manage many images in an in-core way.
	
	\item \emph{Repeatability \& Outliers:} We have to overcome inter-patient variability. When using keypoints, the rate of outliers increases from 30\% for similar patients up to 70\% in case of different patients (see section \ref{sec:A+C}). Two main reasons stand out, both induced by inter-patient variability : (1) match selectivity needs to be decreased to find more matches between differing patients, (2) match consistency cannot be enforced (if three points A, B, C are linked by two matches : AB and BC, A and C must match). As a result, the rate of outliers can exceed the M-estimator breakdown point (50\%), hence the need for a more robust estimator.
	
	\item \emph{Computational complexity: } The analogy with BA is not perfect. BA processes a set of 2D images distributed in a 3D space. As each image overlaps with few other images, BA can be significantly simplified to a band matrix problem. In our case, we need to register 3D images in a 3D space, and the overlap between images is much higher: the images are all located within the human body, but for different patients. This results in a dense matrix problem, illustrated by Figure \ref{fig:sparsity}.
	
\end{itemize}

To overcome these challenges, we introduce in this paper three main contributions:

\begin{itemize}
	
	\item \emph{Keypoint-based hubless registration: }As input data, we use only the keypoints extracted from the images. We still use an abstract common space and half-transforms as in star groupwise registration, but we do not need any central reference data : our optimization is driven only by inter-image registration, as shown in Figure \ref{fig:hubless_groupwise}. Half-transforms are represented by spline pyramids, and we use the graph linking all images together (the complete graph) for the optimization. This compact framework can register 100 images with a memory footprint of only 10GB.
	
	\item \emph{EM-weighting:} To manage a ratio of outliers that can exceed $50\%$, we devise an Expectation-Maximization algorithm which explicitly estimates inlier and outlier distributions. Moreover, experimental results show that the proposed EM-weighting yields better selectivity against outliers than the M-estimator. As a result, EM-weighting allows our algorithm to converge, while the M-estimator failed to provide convergence.
	
	\item \emph{Efficient optimization:} We propose a fast energy function minimization algorithm suited to our nonrigid problem. With this algorithm, our 24-core testing computer registers 20 images (512x512x400 voxels each) in about 10mn on our test machine. A laptop needs about 30mn.
\end{itemize}

\begin{figure}
	\includegraphics[width=\linewidth]{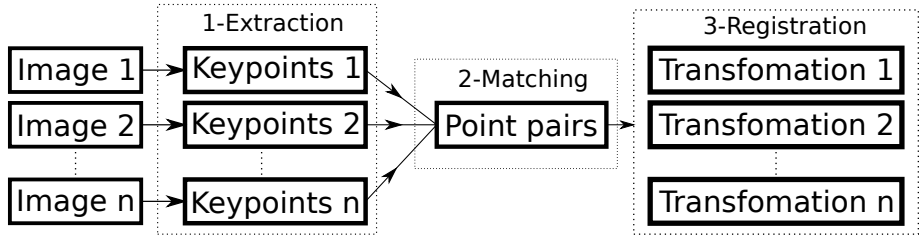}
	\caption{The three main steps in our algorithm: (1) Keypoints are extracted from each image. (2) Matching constructs a list of point pairs from the point sets. (3) Registration optimizes a transformation for each image in order to minimize point pair distances in the common space.}
	\label{fig:outline}
\end{figure}

\section{Method}
\label{sec:method}
Our algorithm proceeds in three steps : (1) 3D SURF keypoint extraction from all input images, (2) points pairing according to their descriptor, (3) half-transforms optimization by minimizing the distance between paired points, while aiming at discarding outlier influence. Figure \ref{fig:outline} shows a block diagram of the three steps.
\subsection{Half transforms driven by keypoints}

\renewcommand{\arraystretch}{1}
\begin{figure*}
	\centering
		\scalebox{.65}{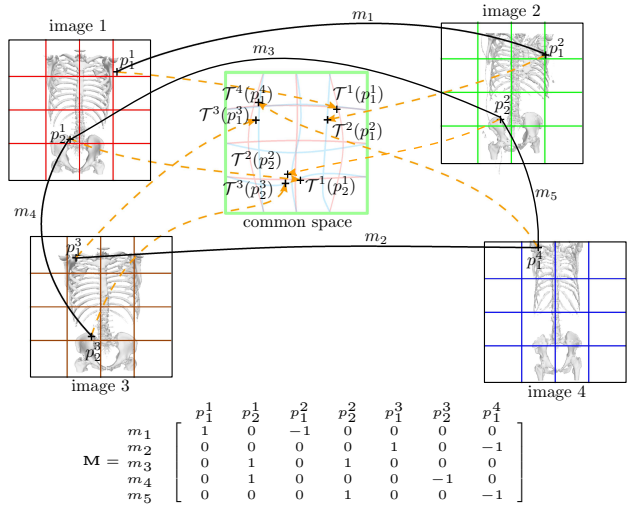}\\
		{\tiny
			$  \matr{M} = 			\kbordermatrix{
				& p^1_1 & p^1_2 & p^2_1 & p^2_2 & p^3_1 & p^3_2 & p^4_1\\
				m_1 	& 1 	& 0 	& -1 	& 0 	& 0	& 0	& 0	\\
				m_2 		& 0 	& 0 	& 0 	& 0 	& 1	& 0	& -1    \\
				m_3 	& 0 	& 1 	& 0 	& 1 	& 0	& 0	& 0     \\
				m_4 	& 0 	& 1 	& 0 	& 0 	& 0	& -1	& 0    \\
				m_5 	& 0 	& 0 	& 0 	& 1 	& 0	& 0	& -1    \\
			}$
		}
	
	\caption{Example with 4 input images. For clarity purposes, only bone structures are displayed. Overprinted on each image $i$: the grid on which lie the splines used to represent the nonrigid transform $\tau^i$. In the middle : the common space, which is used to measure match distances. Similarly to Figure \ref{fig:hubless_groupwise}, matches are drawn in solid black while point transforms are drawn in dashed orange. In this simple example, 7 keypoints were extracted : $\mathcal{P} = \{p^1_1,p^1_2,p^2_1,p^2_2,p^3_1,p^3_2,p^4_1\}$, and 5 matches were found: $\mathcal{M}= \left \{ (p^1_1, p^2_1),  (p^3_1, p^4_1), (p^1_2, p^2_2), (p^1_2, p^3_2), (p^2_2, p^4_1) \right \}$. The match $(p^2_2, p^4_1)$ is an outlier. $\matr{M}$ is the match matrix.}
	\label{fig:halfTransforms}
\end{figure*}

We consider a set $\mathcal{I}$ of $n$ 3D images. In order to get rid of dense computation at the earliest point in the process, we extract for each image $i$ the 3D SURF keypoint set $\mathcal{P}^i = \{p_a^i\}$ as done by \cite{agier2016hubless}. For simplicity, we define $\mathcal{P} = \bigcup \mathcal{P}^i$. The entire algorithm uses physical coordinates (in millimeters), and is invariant to image sampling conditions. We compute the set of matched keypoints $\mathcal{M} = \{(p_a^i, p_b^j)\}$ for all image pairs $(i,j)$, by combining several criteria: feature vector distance, nearest neighbor ratio and Laplacian sign as in \cite{bay2006surf} and scale difference. As explained earlier in the paper, $\mathcal{M}$ contains outliers. Optimizing the set of half transforms $\mathcal{T} = \{\tau^i\}$ is done by minimizing the match $(p_a^i, p_b^j)$ distance-errors : 

\begin{equation}
d(p_a^i, p_b^j) = \|\tau^i(p_a^i) - \tau^j(p_b^j)\|
\label{eq:reproErr}
\end{equation}

This $L_2$ distance is the only measure computed in the abstract common space. We use tensor products of Uniform Cubic B-splines to interpolate the half-transforms $\tau^i$, driven by $c$ control points placed on a rectilinear grid. Each half-transform $\tau^i$ is expressed as: $ \tau^i : p \in \mathbb{R}^3 \mapsto p + \matr{b}(p) \matr{x}^i$, where the $j^{\text{th}}$ element of $\matr{b}(p)$ is the evaluation of the $j^{\text{th}}$ spline basis function at point $p$; $\matr{x}^i$ is the $c \times 3$ spline coefficient matrix which represents the control point displacements (one column for each coordinate).  $\matr{x}^i$ are unknowns we want to optimize for each image.
Figure \ref{fig:halfTransforms} gives an overview of our framework, a simple example with 4 input images, 7  keypoints and 5 matches.

\subsection{Outliers - EM-weighting}

An efficient way of mitigating the presence of outliers in the set of paired points is to use robust statistics such as M-estimators \citep{fox2002robust}. The usual assumption made with M-estimators is that the probability distribution of the distance (\ref{eq:reproErr}) is a combination of one normal distribution (the inlier contribution), and one uniform distribution (the outlier contribution). M-estimators need to estimate inlier variance, the tuning parameter, which provides a way to balance the contribution of each distance $d(p_a^i, p_b^j)$  (\ref{eq:reproErr}) to the optimization criterion. Smaller values of the tuning constant increase robustness to outliers.

In our case, the underlying distributions follow different laws: 
\begin{itemize}
	\item \textbf{Inliers} are norms of Gaussian random vectors of $\mathbb{R}^3$. They follow a $\chi$ distribution with 3 degrees of freedom, also known as Maxwell distribution.
	\item \textbf{Outliers} are distances between random vectors of $\mathbb{R}^3$ following uniform distributions. We propose to approximate their distribution using also a Maxwell distribution. 
\end{itemize}

Then, the probability density function of the distance can be expressed on each image $i$ as:

\begin{equation}
\begin{split}
P^i(d) = r^if(d, s_1^i) + (1-r^i)f(d, s_2^i)
\\
f(d,s) =  \sqrt{\frac{2}{\pi}} \frac{d^2}{s^3} e^{- \frac{d^2}{2s^2}}
\end{split}
\label{eq:Maxwell}
\end{equation}

where $f$ is the Maxwell probability density function, $s_1^i$ and $s_2^i$ are respectively the inlier and outlier scale parameters (the standard deviation of all vector distance coordinates with zero mean) of the two Maxwell laws and $r^i$ is their mixing ratio. Instead of estimating the inlier variance, we directly estimate for each image $i$ the parameters $ \theta^i = (s_1^i, s_2^i, r^i)$ of $P^i(d)$ using the Expectation-Maximization (EM) algorithm \citep{dempster1977maximum}. As we explicitly estimate the Maxwell laws, we can handle more than 50\% of outliers as we do not rely on the median operator. Following Bayes' rule, we can infer for any match distance-error $d$ the probability of belonging to the inlier class $I$ :

\begin{equation}
P(I|d, \theta^i) = \frac{r^if (d, s_1^i )}{r^i.f(d, s_1^i) + (1-r^i)f(d, s_2^i)}
\label{eq:EMestimation}
\end{equation}

\begin{figure}
	\centering
	\includegraphics[width=.4\linewidth]{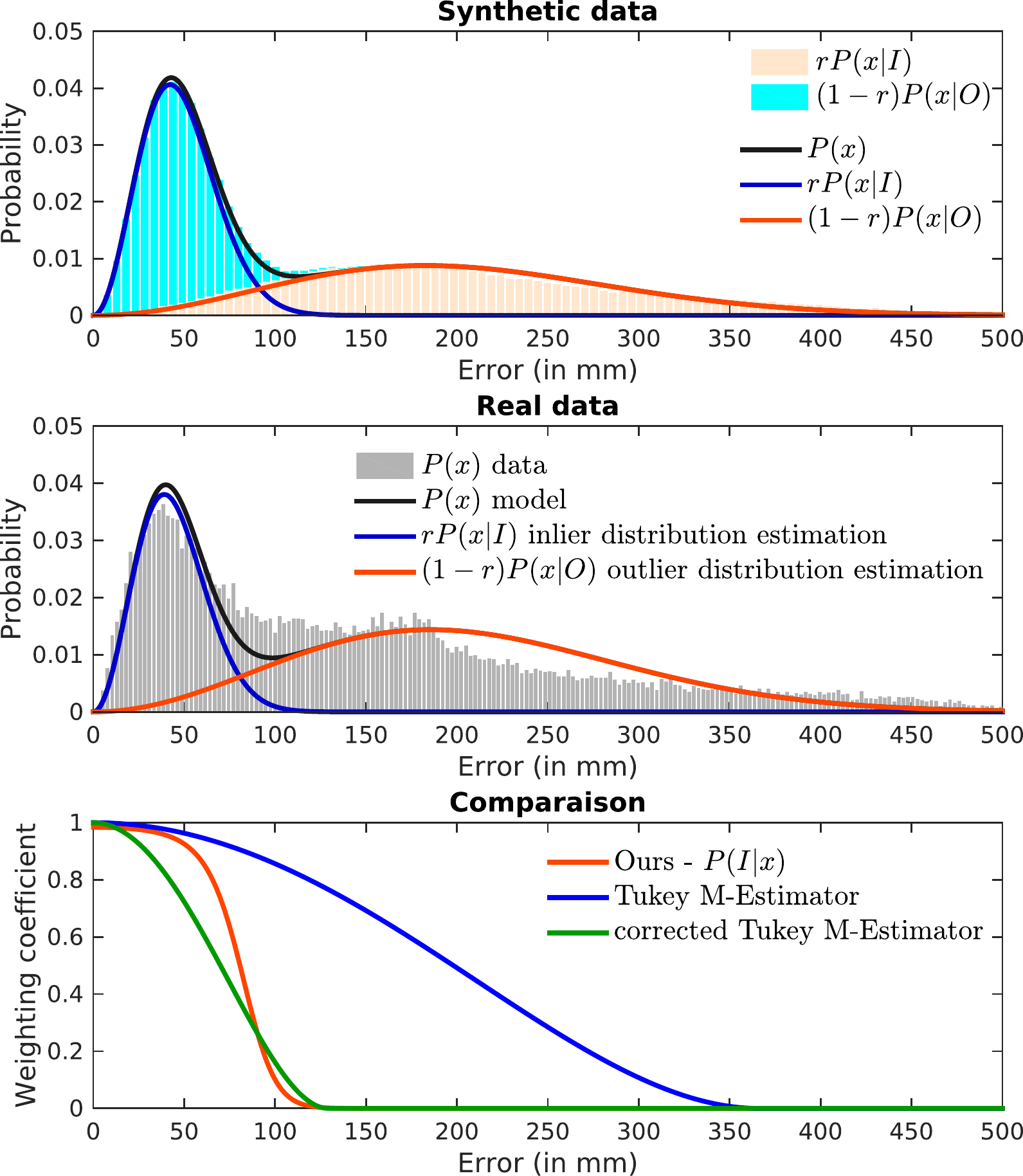}
	\caption{EM-weighting and M-estimators. Top : synthetic distance distribution, with inlier distribution in light blue and outlier distribution in orange. The dark-blue and red curves represent their approximations with Maxwell distributions. Middle : Real world example, with more than 50 \% of outliers. The gray histogram is computed for one image of the VISCERAL data set after registration. The dark-blue and red curves respectively represent the estimation of inlier and outlier distributions with Maxwell distributions. Bottom : Comparison of weighting functions computed with the same real world data. Blue: incorrect standard deviation estimation leads the M-Estimator to include a lot of outliers. Green: M-estimator with corrected standard deviation. Red: our EM-weighting yields better selectivity.}
	\label{fig:distrib}
\end{figure}

This probability can be used as a weighting function to inhibit outlier contributions. But as each match $(p_a^i, p_b^j)$ links two images, two probabilities can actually be computed for each match: $P(I|(p_a^i, p_b^j), \theta^i)$ and $P(I|(p_a^i, p_b^j), \theta^j)$, using respectively image $i$ and image $j$ as statistical context. As a symmetric criterion is preferable, we select the minimum value as a weight for each match:
\begin{equation}
w(p_a^i, p_b^j) = min[P(I|d(p_a^i, p_b^j), \theta^i), P(I|d(p_a^i, p_b^j), \theta^j)]
\label{equ:EM-weighting}
\end{equation}
The minimum is a good heuristic since it increases selectivity against outliers. It has been observed in our experiments that it is more effective to wrongly reject more inliers than wrongly accept more outliers. Figure \ref{fig:distrib} shows example data as well as a comparison between EM-weighting and M-estimators, showing that our approach is more robust and more selective than the M-estimator.

\subsection{Energy minimization}

\subsubsection{Initialization}

Before non-rigid registration, we optimize a 6-parameter transformation (translation + anisotropic scale)  $\lambda^i$ defined for each image $i$ as:
\begin{equation}
\lambda^i : p \mapsto \lambda^i(p) = s^i \circ p + t^i
\end{equation}

Where $s^i$ and $t^i$ are vectors of $\mathbb{R}^3$ representing the anisotropic scale and translation, respectively. The $\circ$ symbol stands for the component-wise Hadamar product. Optimization is carried out using an iterative mean and variance matching algorithm : for each iteration and each image $i$, the scale $s_i$ is updated to $\tilde{s_i}$ so that the component-wise variance $v_a$ of keypoints $p_a^i$ in $i$ gets closer to the variance $v_b$ of their paired keypoints :

\begin{equation}
\tilde{s_i} = ( (v_{bx} / v_{ax})^{\frac{\gamma}{2}}, (v_{by} / v_{ay})^{\frac{\gamma}{2}}, (v_{bz} / v_{az}))^{\frac{\gamma}{2}} )^T
\end{equation}

where $\gamma$ is the update coefficient, lower than 1, and $v^T$ is the transpose of vector $v$.
The translation $t_i$ is updated to $\tilde{t_i}$ to displace in the common space the mean $\overline{p_a}$ of the keypoints $p_a^i$ closer to the mean $\overline{p_b}$ of their paired keypoints:
\begin{equation}
\tilde{t_i} = t_i + \gamma (\overline{p_b} - \overline{p_a}) + \overline{p_a} \circ((1,1,1)^T - \tilde{s_i})
\end{equation}




Note that in order to reduce the influence of outliers, paired points are weighted by our EM-weighting scheme $w(p_a^i, p_b^j)$ \eqref{equ:EM-weighting} during computation of the variances ($v_a$ and $v_b$) and the mean values ($\overline{p_a}$ and $\overline{p_b}$). In our experiments, we carried out 50 iterations with $\gamma=0.5$, and update the weights $w(p_a^i, p_b^j)$ once every 10 iterations.

\subsubsection{Deformable registration}

Once linear registration is computed, we propose to find the best set of B-Spline transforms $\mathcal{T^*}$ by minimizing for each keypoint the mean square distance to its matching points, each match contribution \eqref{eq:reproErr} being weighted according to EM-weighting \eqref{equ:EM-weighting}:

\begin{equation}
\matr{E} (\mathcal{T}) = 
\sum_{\vphantom{p_a^i \in \mathcal{P}^i} i=1}^n
\sum_{p_a^i \in \mathcal{P}^i} 
\left(\frac{1}{ \lvert \mathcal{N}(p_a^i) \rvert } \sum_{p_b^j \in \mathcal{N}(p_a^i)} w(p_a^i, p_b^j)d(p_a^i, p_b^j)^2\right) \\
\label{eq:costFunction}
\end{equation}
where $\lvert \mathcal{N}(p_a^i) \rvert$ is the number of points matching with $p_a^i$.

However, $n-1$ transforms are enough to register $n$ volumes whereas there are $n$ unknown transforms $\tau^i$ in our problem. This results in an underdetermined problem: convergence is not guaranteed and the transforms can be subject to global drift. Following \cite{wu2012feature}, we remove the degree of freedom by adding an extra constraint to the control point displacements:

\begin{equation}
\mathcal{T}^* = \min_{\mathcal{T}} \matr{E}(\mathcal{T}) ~~~  s.t. \sum_{i=1}^n \matr{x}^i =\matr{0}
\label{eq:minConstraint}
\end{equation}

By traversing the set of matches $\mathcal{M}$, The energy function (\ref{eq:costFunction}) can be rewritten as: 

\begin{equation}
\matr{E} (\mathcal{T}) =  \sum_{(p_a^i, p_b^j) \in \mathcal{M}} \left (s(p_a^i, p_b^j)d(p_a^i, p_b^j) \right ) ^2 \\
\label{eq:costFunctionTra}
\end{equation}

where $s(p_a^i, p_b^j)$ is defined by :
\begin{equation}
s(p_a^i, p_b^j) = \sqrt{ w(p_a^i, p_b^j)\left(\frac{1}{\lvert \mathcal{N}(p_{a\vphantom{b}}^{i\vphantom{j}}) \rvert}+\frac{1}{\lvert \mathcal{N}(p_b^j) \rvert}\right)}
\label{eq:matrixS}
\end{equation}

Equation (\ref{eq:costFunctionTra}) can be expressed in matrix notation :
\begin{equation}
\begin{split}
\matr{A} = \matr{S}\matr{M}\left(\matr{P}+\matr{B}\matr{X}\right)\\
\matr{E}(X) = tr(\matr{A}^t\matr{A})
\end{split}
\label{eq:matrixCostFunc}
\end{equation}

where $\matr{S}$ is the $|\mathcal{M}| \times |\mathcal{M}|$ diagonal matrix of all $s(p_a^i, p_b^j)$. 

$\matr{P}$ is the  $|\mathcal{P}| \times 3$ matrix of keypoint locations.

$\matr{M}$ is the sparse $ |\mathcal{M}| \times |\mathcal{P}| $ match matrix. Each line of $ \matr{M}$ represents a match between two keypoints, and contains exactly two nonzero values : one $1$ and one $-1$ in the columns corresponding to the two matching keypoints. $ \matr{M}$ can be seen as the incidence matrix of the match directed graph where vertices are keypoints and edges are matches with arbitrary orientation. An example of $ \matr{M}$ is shown by Figure \ref{fig:halfTransforms}.

$\matr{B}$ is the $ |\mathcal{P}| \times nc $ spline matrix. $\matr{B}$ is a sparse non-square block diagonal matrix, where each non-zero block $\matr{B}^i$ is the $|\mathcal{P}^i| \times c $ spline matrix for image $i$ (the stacking of all $\matr{b}(p_a^i)$).

$\matr{X}$ is the $ nc \times 3 $ stacking of all matrices $\matr{x}^i$.

To minimize equation \eqref{eq:costFunctionTra}, inspired by the Iteratively Reweighted Least Squares method, we use a gradient descent algorithm and we update the parameters $\theta^i$ every ten iterations. The gradient of equation (\ref{eq:matrixCostFunc}) is : 

\begin{equation}
\nabla \matr{E} = 2\matr{B}^t\matr{M}^t\matr{S}^2\matr{M}\left(\matr{P} + \matr{B}\matr{X}\right)
\label{equ:gradient}
\end{equation}

\begin{algorithm}[t]
	\caption{One level of gradient descent for deformable registration. $\alpha$ is a small positive coefficient, $nIterations$ is the number of iterations.}
	\label{algo}
	\begin{algorithmic} 

	\STATE $\widetilde{\matr{X}_0} = \matr{0} $
	\STATE  $k \leftarrow 0$
	\WHILE{$k < nIterations$}
	\IF{$k \mod 10 = 0$}
		\STATE Compute the parameters $\theta^i$
	\ENDIF

	\STATE Compute $\matr{S}_{k}$ from $\theta^i$
	\STATE 	$ \matr{X}_{k+1} \leftarrow \widetilde{\matr{X}}_{k} - \alpha 2\matr{B}^t\matr{M}^t\matr{S}^2_{k}\matr{M}\left(\matr{P} +\matr{B}\widetilde{\matr{X}}_{k}\right)$

	\STATE Split $\matr{X}_{k+1}$ into $n$ coefficient matrices $\matr{x}_{k+1}^i$,

	\STATE $\widetilde{\matr{x}}_{k+1}^i \leftarrow \matr{x}_{k+1}^i - \frac{1}{n} \sum_{j=1}^n{\matr{x}_{k+1}^j}$

	\STATE Construct $\widetilde{\matr{X}}_{k+1}$ by stacking the matrices $\widetilde{\matr{x}}_{k+1}^i$
	\STATE $k \leftarrow k+1$
	\ENDWHILE

	\end{algorithmic}
\end{algorithm}

After each iteration, we shift the mean of the spline coefficients in order to satisfy the constraint in \eqref{eq:minConstraint}. Note that an interesting byproduct of this step is the iterative removal of spatial bias in the common space \citep{guimond00}. Our gradient descent approach is summarized by algorithm \ref{algo}.

We use a hierarchical approach, beginning with a coarse spline grid and incrementally doubling the grid density. This improves the speed of our approach and keeps the solution close to the global minimum, which is required as we use EM-weighting. Note that we achieve a significant simplification of the gradient computation \eqref{equ:gradient}, by taking into account the fact that the matrices $\matr{M}$ and $\matr{B}$ are sparse.

\subsection{Guaranteed diffeomorphism}

To guarantee that FROG outputs diffeomorphic half-transforms, we use a method similar to \cite{rueckert2006}: A sufficient condition to guarantee that a B-Spline based transformation is diffeomorphic is to check that for each direction (x, y and z), the displacement $d$ of each control point is below $0.4*g$ where $g$ is the initial distance between control points (the grid spacing). Then, for a given resolution level, instead of using a single transformation which may not be diffeomorphic due to large control point displacements, we use a composition of several transformations, each respecting the diffeomorphism constraint. As the composition of diffeomorphisms is a diffeomorphism, the resulting composition is also a diffeomorphism.

For a given resolution level, optimization begins with a single grid for each half-transform. After each iteration, we check that the diffeomorphism constraint is fulfilled for all half-transforms. When the constraint is not satisfied, the iteration is canceled, current half-transforms are frozen, and iterations are resumed with a new set of half-transforms which will be composed with the previous ones. Subsequent iterations may also break the diffeomorphism condition, hence several transformations can be created for a single resolution level. As a result, for a given level, each half-transform is the concatenation of several transformations with limited displacements, which guarantees diffeomorphism. In our experiments on VISCERAL group A, with 3 levels of resolution, the number of concatenated transformations for each level was respectively 5, 8 and 14. Hence, instead of three grids (one for each level), our approach used a total of 27 grids per image. In our experiments, using a single grid for each resolution level yields half-transforms with a few percentage of voxels with negative Jacobian determinant, while using grid compositions yields no negative Jacobian determinant.

\section{Results}
\label{sec:results}

\subsection{Data sets and evaluation}
We evaluate our approach with the VISCERAL database \citep{Langs2012visceral} which is composed of three groups of images (dimensions around $512 \times 512 \times 400$, spacing around $0.7 \times 0.7 \times 1.5$mm, 32 bits floats):
\begin{itemize}
	\item A: 20 volumes (thorax and abdomen) of contrast enhanced (via contrast agent injection) CT scans. For these volumes, experts have located anatomical landmarks (up to 45 per volume, between 41 and 42 on average).
	\item B: 20 volumes (whole body) of CT scans. For these volumes, experts have located anatomical landmarks (up to 53 per volume, between 52 and 53 on average).
	\item C: 63 volumes (thorax and abdomen) of contrast enhanced CT scans, not annotated. The VISCERAL consortium refers to these volumes as the silver corpus.
\end{itemize}

With these three different groups (A, B and C), we are able to carry out several groupwise registrations with different scenarios: only A, only B, A and C or the three groups together, while observing the effects of these cases on the landmarks registration accuracy. To measure the quality of a given groupwise registration, we first project all reference landmarks in the common space using the transforms $\tau^i$. Then, for each landmark category (Clavicle L., Clavicle R. etc...), we compute a mean position $\overline{p}$. The quality criterion for this category is then defined as the average distance to $\overline{p}$. We also compute the global maximum individual distance to $\overline{p}$ as a robustness criterion. Note that reference landmarks are only present for images in groups A and B. As a consequence, when registering groups A and C together, the criterion is restricted to images of group A. Note that landmarks are never used by any of the tested algorithm during registration. Experiments were carried out with a 24-core computer (48 logical cores with hyperthreading) with 128 GB of RAM.

We have compared our FROG approach with star-groupwise voxel-based methods NiftyReg \citep{modat2008, modat2014} and ANTs \citep{Avants2008ants}. As the test data contains high resolution images, other dense methods could not be applied. As an example, Elastix \citep{klein10} and 
GLIRT \citep{wu2012feature} require to store all images in-core, and GRAM \citep{hamm2010} exhibits a computational complexity which is too high for our data sets (see section \ref{sec:registration}).

\subsection{Implementation and settings}
We have implementation FROG in C++, and its code source is publicly available at \href{https://github.com/valette/frog}{https://github.com/valette/frog}.

In the complexity analysis, the dominant terms are $| \mathcal{I}| v \log{v}$ for keypoint extraction, $| \mathcal{P}|^2$ for match computation and $|\mathcal{I}||\mathcal{P}| c \log{c}$ for optimization, where $|\mathcal{I}|$ is the number of volumes, $v$ is the number of voxels for each volume, $|\mathcal{P}|$ is the total number of extracted keypoints, and $c$ is the number of control points at the highest resolution.

Our pyramid of splines contains three levels, with a grid step of $200mm$, $100mm$, $50mm$ respectively. For each volume, we keep up to 20000 keypoints (see \cite{agier2016hubless}). For each resolution level, we compute 200 gradient descent iterations, with $\alpha$ set to $0.02$.

\subsection{Comparison with NiftyReg and ANTs}

With NiftyReg, we have used the \emph{groupwise\_niftyreg\_run.sh} script. This script computes a star-groupwise registration (see Figure \ref{fig:star_groupwise}) and generates a template model from all input volumes. It alternates between mean reference computation and pairwise registration of all images with this reference. By default, 15 iterations of groupwise registration are applied: the first 5 iterations use rigid registration, the 10 last iterations use nonrigid registration.

With ANTs; we have used the \emph{antsMultivariateTemplateConstruction2.sh} script. This script computes a star-groupwise registration (see Figure \ref{fig:star_groupwise}) and generates a template model from all input volumes. It alternates between mean reference computation and pairwise registration of all images with this reference. By default, 5 iterations of groupwise registration are applied: the first is a rigid registration, the 4 others being nonrigid. The best results with ANTs have been obtained using mutual information as registration criterion.

\subsubsection{Group A}
Detailed results are given by Tables \ref{tab:ours_vs_ants} and \ref{tab:ours_vs_ants_global}.

ANTs processes the 20 volumes in 62.5 hours, with a  mean registration distance of 15.5 mm. To further increase the speed of ANTs, we switched the registration subsampling factor $sf$ from 1 to 2 and 4 (i.e. setting the shrinking factor parameter to 12x6x4x2 and 20x12x6x4, respectively). Setting $sf=2$ increases both speed and registration quality: the 20 volumes are registered in 13.8 hours with a mean distance of 11.3 mm. In that case, there is no wrongly registered volume. Setting $sf=4$ further increases computing speed but decreases output quality: the 20 volumes are registered in 9.4 hours with a mean distance of 14.0 mm. These results are presented in Table \ref{tab:ours_vs_ants_global}, where ANTS-S2 and and ANTS-S4 refer to setting $sf$ to 2 and 4, respectively.

NiftyReg registered group A in 67 hours on our test machine, yielding an average landmark distance of $9.6 \pm 7.5 mm$. 

\begin{figure}
	\centering
	\includegraphics[width=.6\linewidth]{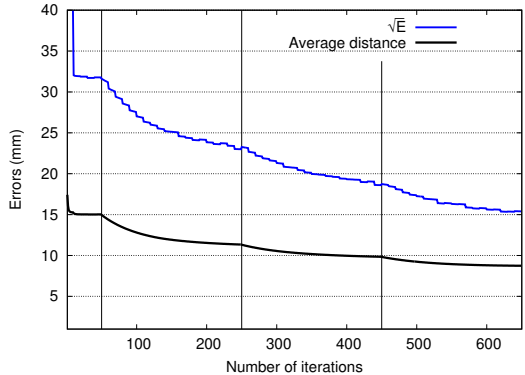}
	\caption{Convergence of the minimized criterion and mean distance when registering group A. Distances are in mm. The blue curve is the square root of the criterion given by Equation \eqref{eq:costFunction} which reflects the average distance between matched keypoints. The black curve represents the global mean landmark distances. For the first 50 iterations, we use an anisotropic scale+translation model. For the next iterations, we use B-splines over 3 different resolution levels, 200 iterations for each level.}
	
	\label{fig:convergence}
\end{figure}

FROG was able to successfully register all volumes of group A, while being much faster than NiftyReg and ANTs. Registration took 10mn on our test machine, and less than 30mn on a 4-core laptop. The two-tailed Welch's t-test shows that our approach (Table \ref{tab:ours_vs_ants} third column) achieves an average landmark distance ($8.7\pm 7.2 mm$) significantly lower to that of NiftyReg ($9.6\pm9.6mm$, $p-value=0.006$) and ANTs ($11.3 \pm 7.9mm$, $p-value =3.3.10^{-18}$). We have also added experiments to table \ref{tab:ours_vs_ants_global}, varying the number of kept keypoints for each image to respectively 40k, 30k, 10k, 5k and 2.5k. Note that as expected, accuracy increases with the number of extracted keypoints, but saturates above 30k points per volume. Hence we have fixed the number of kept points to 20k. Figure \ref{fig:convergence} shows the convergence curve for $\sqrt{E}$ (see equation \ref{eq:costFunction}) in blue and the mean landmarks distance during optimization in black. All distances are in mm.

\begin{table}[]
	\caption{ Registering Group A : comparison of landmark average distances (in mm) between our approach (FROG), NiftyReg and ANTs.}
	\label{tab:ours_vs_ants}
	\small
	\centering
	\begin{tabular}{l|r|r|r}
		\hline 
		\bfseries Landmark & \bfseries ANTs & \bfseries NiftyReg & \bfseries FROG\\
		\hline
Clavicle L & $ 9.7 \pm 4.6 $ & $ \bm{6.8 \pm 5.1} $ & $ 8.6 \pm 2.4 $\\
Clavicle R & $ 9.3 \pm 4.4 $ & $ \bm{7.1 \pm 3.5} $ & $ 7.6 \pm 3.4 $\\
Tubercul. L & $ 11.2 \pm 5.4 $ & $ 11.6 \pm 6.1 $ & $ \bm{9.3 \pm 2.4} $\\
Tubercul. R & $ 10.9 \pm 5.5 $ & $ 11.7 \pm 5.7 $ & $ \bm{9.6 \pm 5.2} $\\
C6 & $ \bm{3.5 \pm 2.6} $ & $ 4.1 \pm 1.7 $ & $ 5.3 \pm 2.3 $\\
C7 & $ \bm{5.0 \pm 3.3} $ & $ 7.6 \pm 3.8 $ & $ 6.7 \pm 4.3 $\\
Th1 & $ \bm{4.5 \pm 3.0} $ & $ 7.6 \pm 4.2 $ & $ 6.7 \pm 5.6 $\\
Th2 & $ \bm{4.9 \pm 3.4} $ & $ 6.6 \pm 3.7 $ & $ 6.0 \pm 4.9 $\\
Th3 & $ 6.1 \pm 4.2 $ & $ 6.9 \pm 3.8 $ & $ \bm{5.7 \pm 4.4} $\\
Th4 & $ 7.6 \pm 4.3 $ & $ 7.9 \pm 3.4 $ & $ \bm{5.9 \pm 4.6} $\\
Th5 & $ 8.6 \pm 3.6 $ & $ 8.4 \pm 3.0 $ & $ \bm{6.8 \pm 4.6} $\\
Th6 & $ 9.6 \pm 3.4 $ & $ 9.9 \pm 3.5 $ & $ \bm{7.8 \pm 5.1} $\\
Th7 & $ 10.2 \pm 3.4 $ & $ 10.2 \pm 3.8 $ & $ \bm{8.4 \pm 6.1} $\\
Th8 & $ 10.3 \pm 3.9 $ & $ 11.0 \pm 4.7 $ & $ \bm{8.9 \pm 8.2} $\\
Th9 & $ \bm{10.8 \pm 4.9} $ & $ \bm{10.8 \pm 3.8} $ & $ \bm{10.8 \pm 9.8} $\\
Th10 & $ \bm{11.3 \pm 5.4} $ & $ \bm{11.3 \pm 3.6} $ & $ 11.7 \pm 11.1 $\\
Th11 & $ 12.2 \pm 5.9 $ & $ \bm{11.8 \pm 5.9} $ & $ 13.0 \pm 10.3 $\\
Th12 & $ 13.4 \pm 6.2 $ & $ \bm{11.8 \pm 7.3} $ & $ 13.5 \pm 9.6 $\\
L1 & $ 13.3 \pm 6.6 $ & $ \bm{11.9 \pm 7.6} $ & $ 12.7 \pm 9.5 $\\
L2 & $ 12.8 \pm 6.6 $ & $ 11.6 \pm 8.3 $ & $ \bm{10.6 \pm 7.8} $\\
L3 & $ 13.1 \pm 7.1 $ & $ \bm{11.0 \pm 8.3} $ & $ 11.6 \pm 6.5 $\\
L4 & $ 12.8 \pm 7.2 $ & $ \bm{10.0 \pm 7.2} $ & $ 10.1 \pm 6.0 $\\
L5 & $ 11.4 \pm 6.7 $ & $ 8.4 \pm 6.7 $ & $ \bm{7.8 \pm 6.4} $\\
Sternoclav. L & $ 7.9 \pm 5.5 $ & $ 8.5 \pm 5.7 $ & $ \bm{5.4 \pm 4.7} $\\
Sternoclav. R & $ 6.6 \pm 3.9 $ & $ 5.0 \pm 4.4 $ & $ \bm{4.9 \pm 3.3} $\\
Aortic arch & $ \bm{8.4 \pm 3.2} $ & $ 9.3 \pm 3.7 $ & $ 9.1 \pm 3.9 $\\
Trachea bif. & $ \bm{4.5 \pm 3.8} $ & $ \bm{4.5 \pm 4.0} $ & $ 4.6 \pm 2.0 $\\
Bronchus L & $ 7.5 \pm 5.5 $ & $ 8.2 \pm 8.6 $ & $ \bm{6.8 \pm 3.6} $\\
Bronchus R & $ 5.9 \pm 4.0 $ & $ 6.4 \pm 7.8 $ & $ \bm{4.5 \pm 2.0} $\\
Coronaria & $ 9.1 \pm 6.4 $ & $ 10.1 \pm 10.3 $ & $ \bm{7.5 \pm 2.6} $\\
Aortic valve & $ 11.8 \pm 7.2 $ & $ 13.7 \pm 10.1 $ & $ \bm{9.4 \pm 3.8} $\\
Xyphoideus & $ 16.6 \pm 7.8 $ & $ 15.5 \pm 9.6 $ & $ \bm{13.5 \pm 9.8} $\\
Renal pelvis L & $ 17.9 \pm 9.2 $ & $ 10.9 \pm 10.6 $ & $ \bm{7.3 \pm 3.3} $\\
Renal pelvis R & $ 20.6 \pm 15.1 $ & $ 16.0 \pm 13.4 $ & $ \bm{11.9 \pm 13.3} $\\
Crista iliaca L & $ 9.9 \pm 6.0 $ & $ \bm{9.4 \pm 6.1} $ & $ 9.7 \pm 6.0 $\\
Crista iliaca R & $ 10.4 \pm 8.8 $ & $ 10.4 \pm 8.7 $ & $ \bm{9.5 \pm 7.3} $\\
Aorta bif. & $ 15.6 \pm 10.5 $ & $ 11.2 \pm 7.3 $ & $ \bm{9.4 \pm 5.4} $\\
VCI bif. & $ 12.6 \pm 6.5 $ & $ 9.1 \pm 5.2 $ & $ \bm{8.2 \pm 4.3} $\\
Troch. maj. L & $ 19.3 \pm 10.8 $ & $ \bm{15.8 \pm 9.1} $ & $ 15.9 \pm 8.8 $\\
Troch. maj. R & $ 18.5 \pm 9.1 $ & $ 16.5 \pm 9.7 $ & $ \bm{14.9 \pm 8.4} $\\
Ischiadicum L & $ 11.8 \pm 6.4 $ & $ \bm{4.3 \pm 3.5} $ & $ 4.6 \pm 3.2 $\\
Ischiadicum R & $ 10.9 \pm 6.3 $ & $ \bm{4.8 \pm 2.9} $ & $ 5.7 \pm 3.0 $\\
Symphysis & $ 15.8 \pm 9.9 $ & $ 10.7 \pm 8.5 $ & $ \bm{10.2 \pm 8.0} $\\
Troch. min. L & $ 15.2 \pm 7.7 $ & $ \bm{3.7 \pm 2.0} $ & $ 5.3 \pm 1.9 $\\
Troch. min. R & $ 16.6 \pm 5.8 $ & $ \bm{4.1 \pm 2.7} $ & $ 6.1 \pm 3.0 $\\
		Mean & $11.3 \pm 7.9$ & $9.6 \pm 7.5$ & $\bm{8.7 \pm 7.2}$ \\
		Maximum & $70.6$ & $\bm{57.5}$ &  $65.4$\\ 
		\hline 
		Time (h)& $13.8$ & $67.0$ & $\bm{0.2}$\\
	\end{tabular}


\end{table}

\begin{figure}[t]
	\centering
	\subfigure[Niftyreg]{\includegraphics[width=0.4\linewidth ]{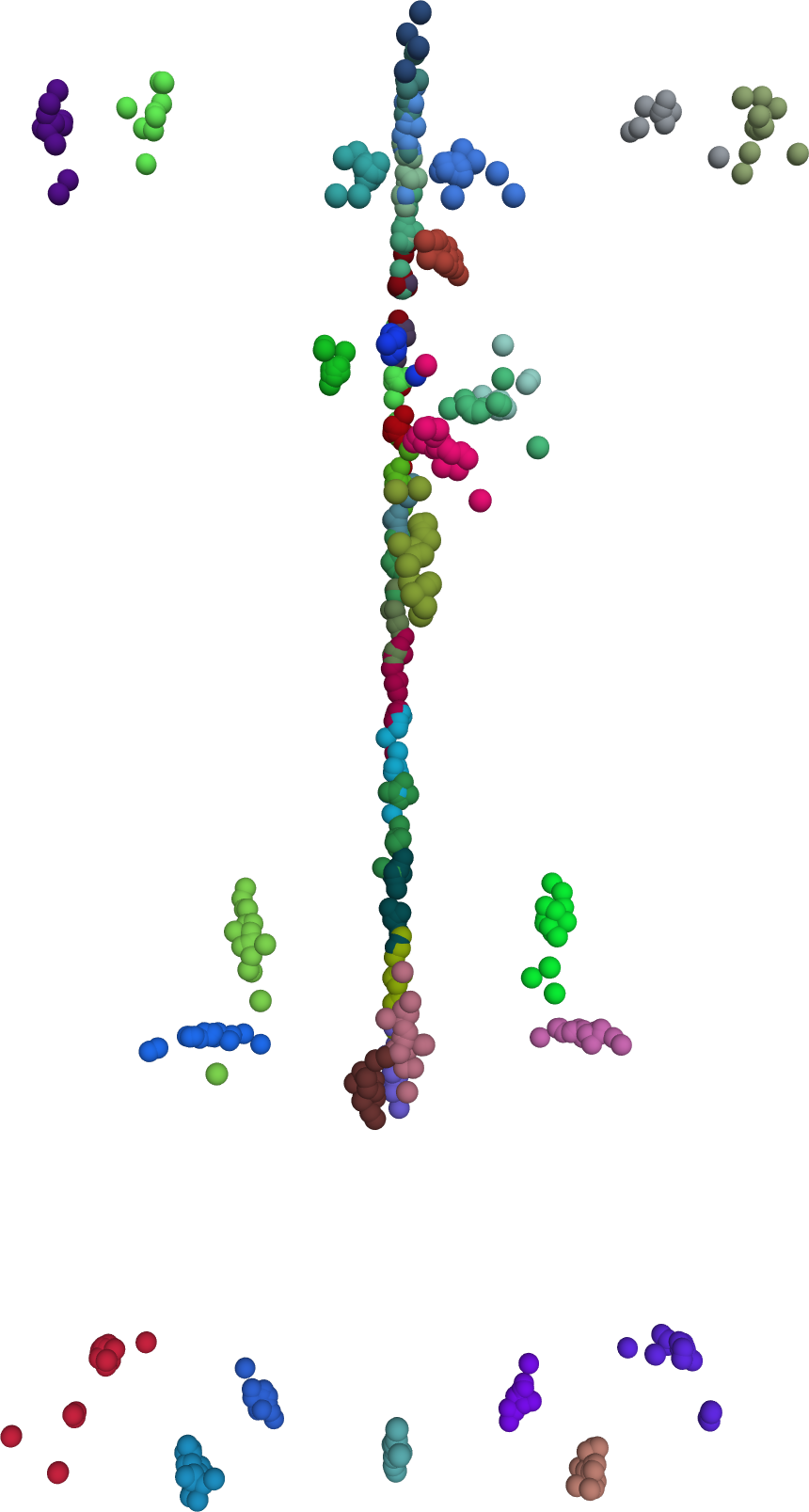}}
	\hfill
	\subfigure[FROG]{\includegraphics[width=0.39\linewidth ]{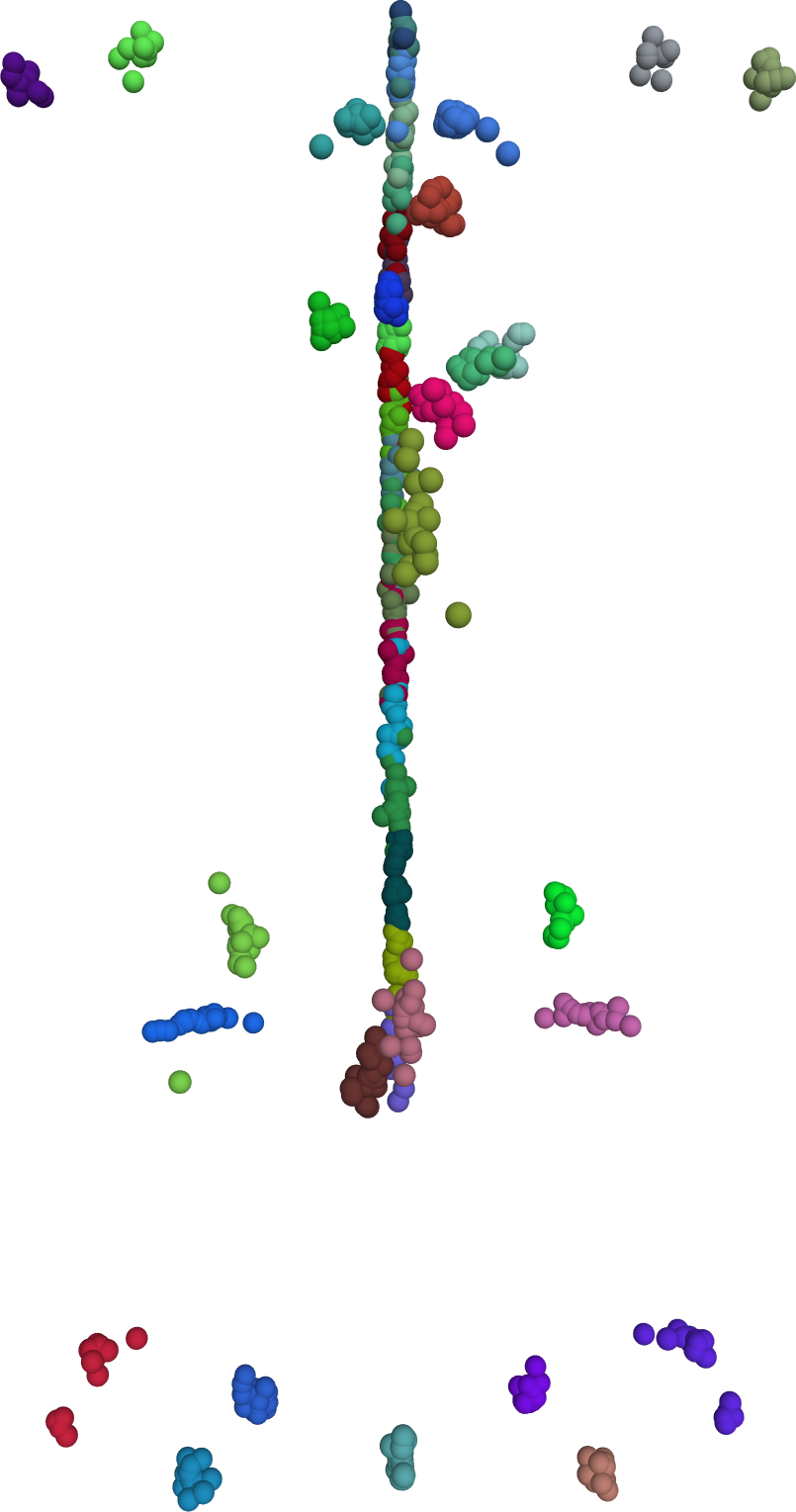}}
	\caption{Comparison between our FROG approach and NiftyReg on VISCERAL group A: landmarks transformed in the common space.}
	\label{fig:landmarks}
\end{figure}

\begin{figure}[]
	\centering
	\includegraphics[width=.7\linewidth, trim={2cm 8.5cm 2cm 8.5cm}, clip]{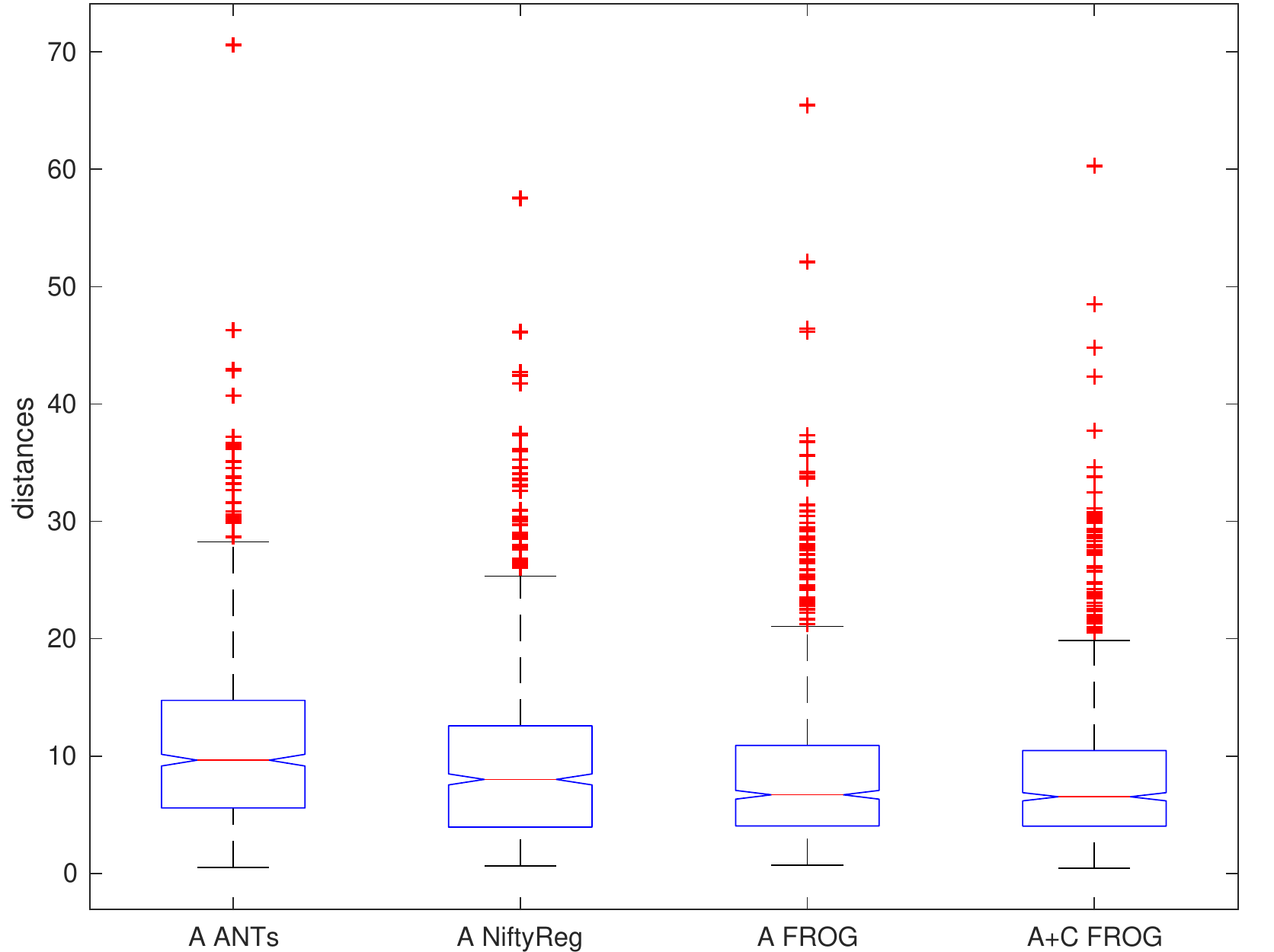}
	\caption{Landmark distances on the set of reference landmarks transformed into the common space. From left to Right : ANTs (group A), NiftyReg (group A), FROG (group A), FROG (groups A+C). Distances are in mm.}
	\label{fig:boxplots}
\end{figure}

Figure \ref{fig:landmarks} compares reference landmarks transformed in the common space, for NiftyReg and our approach. Figure \ref{fig:boxplots} summarizes the statistics with boxplots of landmark errors obtained with ANTs (group A), NiftyReg (group A), FROG (group A) and FROG (groups A+C). On each plot, the central mark is the median, the edges of the box are the $25^{th}$ (Q1) and $75^{th}$ percentiles (Q3), the whiskers ($W=1.5$) extend to the most extreme data points not considered to be outliers, and the outliers are plotted individually if they are larger than $Q3+W(Q3-Q1)$. $W=1.5$ corresponds approximately to +/- 2.7 standard deviation and 99.3 coverage if the data is normally distributed. Notches draw comparison intervals. Two medians are significantly different at the 5\% level if their intervals do not overlap. The interval endpoints are the extremes of the notches. If the notches in the box plot do not overlap, we can conclude with 95\% confidence that the true medians do differ.

\subsubsection{Group B}

\begin{figure}
	\centering
	\subfigure[ANTs]{\includegraphics[width=0.19\linewidth]{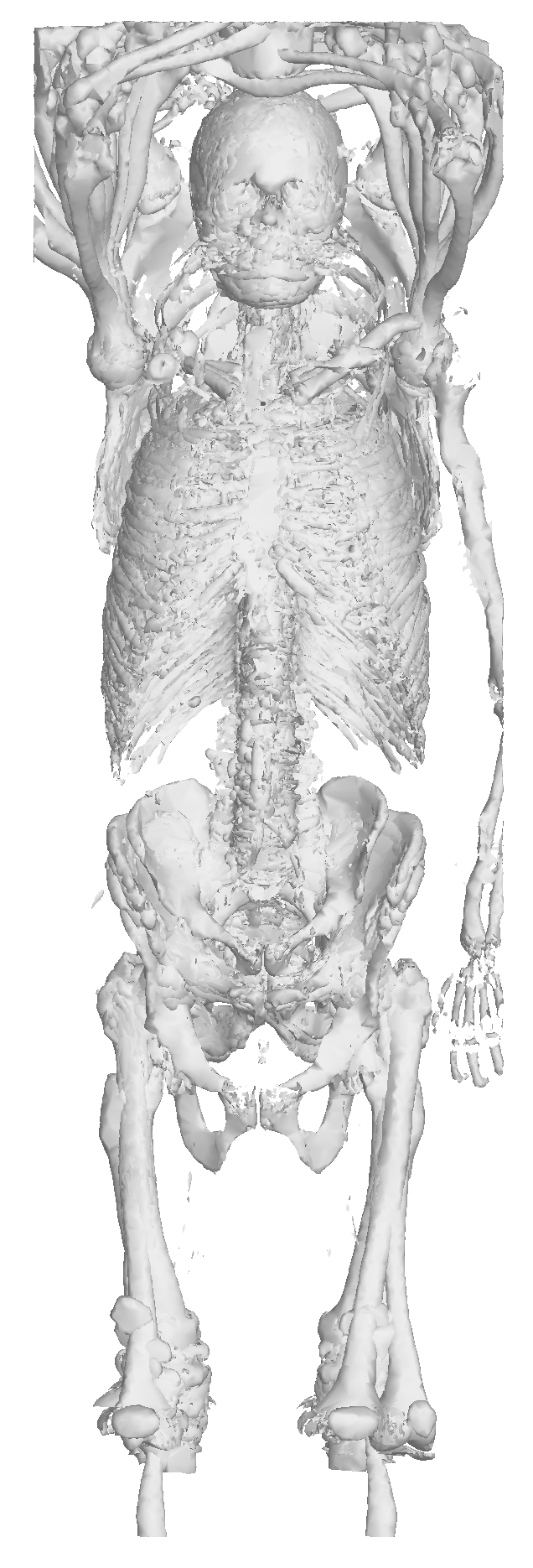}}
	\hspace{10mm}
	\subfigure[NiftyReg]{\includegraphics[width=0.185\linewidth]{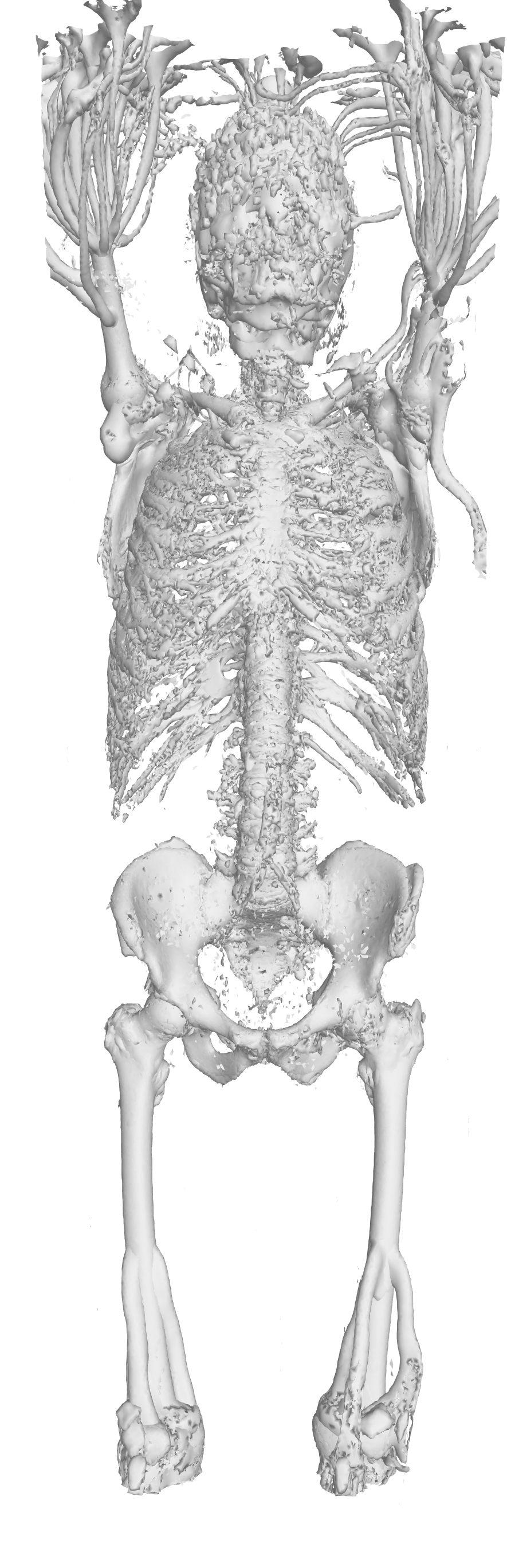}}
	\hspace{10mm}
	\subfigure[FROG]{\label{fig:wb_ours}\includegraphics[width=0.18\linewidth]{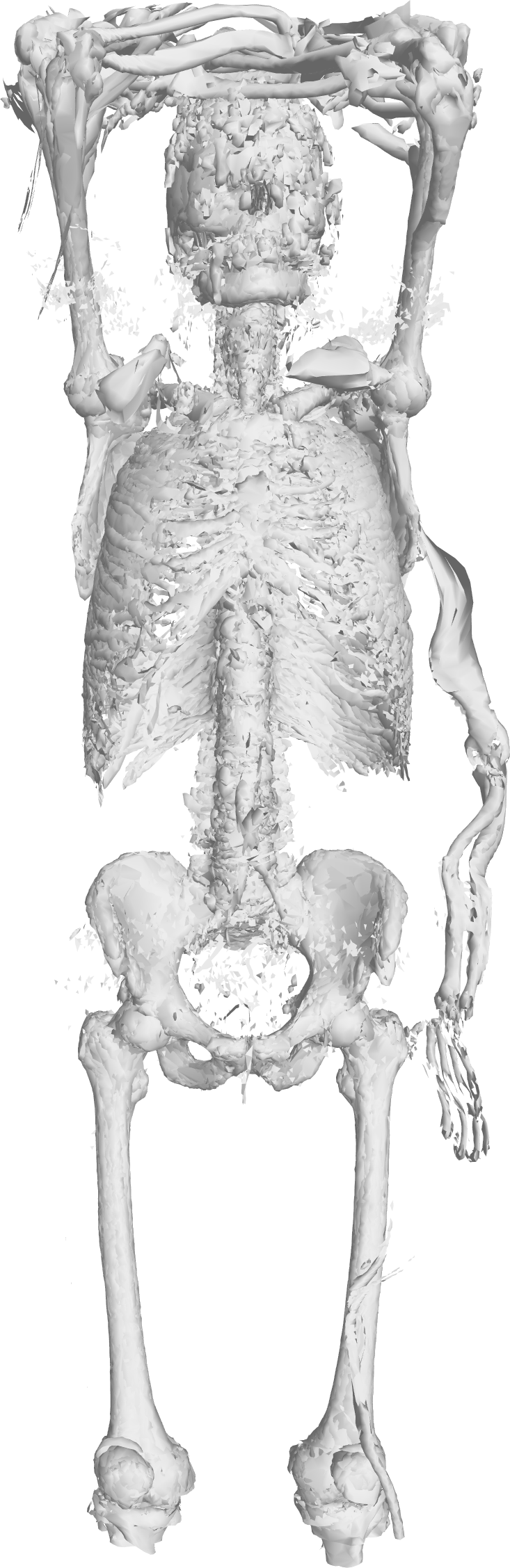}}
	\caption{Bone structures extracted from a registered group of 20 whole-body CT volumes of the VISCERAL group B. Groupwise registration has been carried out with (a) ANTs, (b) NiftyReg and (c) our algorithm. Our algorithm exhibits a better overall robustness to variability.}
	\label{fig:wb}
\end{figure}

Figure \ref{fig:wb} compares groupwise registration of group B. We have extracted bone structures from each volume and superposed them in the common space, using (a) ANTs, (b) NiftyReg and (c) FROG. NiftyReg yields a better result on the torso and hip regions, but still lacks robustness for the legs. Our approach is much more robust to variability, as most of its error is concentrated around the patient arms, which are very challenging regions due to different shoulder poses. The average landmark distance (reported in the second block of Table \ref{tab:ours_vs_ants_global}) obtained with our approach is $9.2 \pm 8.3 mm$, significantly lower than NiftyReg ($12.8 \pm 11.6mm$) and ANTs ($17.9 \pm 27.3 mm$). Note that landmark distances are larger than the ones obtained on group A, for all algorithms (Ours, NiftyReg, ANTs). One reason is that the images of group B were obtained without any contrast agent. As a consequence, soft tissue is much less visible on group B, which could explain lower accuracy.

\subsubsection{Groups A and C}
\label{sec:A+C}

When registering groups A and C together, ANTs yields a mean distance of 44.7 mm. One explanation is that computing the average image from 83 full body scans with a high variability can lead to a very blurred image when registration is not perfect (top right image in Figure \ref{fig:average}). As a consequence, registration against a blurry reference image degenerates.

Registering groups A and C together took 324 hours with NiftyReg and has resulted in an average landmark distance of $9.3 \pm 9.3 mm$.

Our algorithm registers groups A and C within 1.3 hours and yields an average distance of $8.5 \pm 6.9 mm$, statistically better than NiftyReg ($9.3 \pm 9.3 mm$, $p-value=0.0045$) and significantly better than ANTs ($44.7 \pm 31.6 mm$). We have estimated a rate of outliers of 70\% when registering groups A and C together. Note that our previous work based on rigid groupwise registration \citep{agier2016hubless} yields an average error of 31.7mm.

\begin{table}
	\caption{Comparison between our approach, ANTs and NiftyReg on the VISCERAL data set. Our approach extracts 20k points per volume, except for lines 2 to 6 where the number of points vary between 40k and 2.5k}.
	\label{tab:ours_vs_ants_global}
	\centering
	\small
	
	\begin{tabular}{|r|r|r|r|S[table-format=3.1,table-figures-uncertainty=3]|r|}
		\bf Groups& \bf N & \bf Algorithm & \bf Max $d$ (mm) & \bf \hspace{3mm}  $\overline{d}$ (mm) & \bf Time (h)\\
		\hline
		A & 20 & FROG & 65.4  & $\bm{ 8.7 \pm 7.2 }$& $\bm{0.2}$ \\
		A & 20 & FROG (40k) & 67.5  & ${8.7 \pm 7.1}$ & 0.4 \\
		A & 20 & FROG (30k) & 65.9  & ${8.7 \pm 7.1}$ & 0.2 \\
		A & 20 & FROG (10k) & 71.6  & ${9.2 \pm 7.6}$ & 0.1 \\
		A & 20 & FROG (5k) & 63.9  & ${9.9 \pm 8.1 }$&  0.1\\
		A & 20 & FROG (2.5k) & 68.9  & ${10.9 \pm 8.7 }$& 0.1 \\
		A & 20 & FROG (SIFT) & 70.9  & ${11.6 \pm 9.7}$ & 4.2 \\
		A & 20 & ANTs-S4 & 152.7  & 14.0 \pm 17.0  & 9.4\\
		A & 20 & ANTs-S2 & 70.6  & 11.3 \pm 7.9  & 13.8\\
		A & 20 & ANTs &  150.2 &  15.5 \pm 16.7 & 62.5\\
		A & 20 & NiftyReg&  $\bm{57.5}$ &  9.6 \pm 7.5 & 67.0\\
		\hline
		B & 20 & FROG &$ \bm{64.9}$  & $\bm{9.2 \pm 8.3} $& $\bm{0.5}$\\
		B & 20 &  ANTs-S2 & 231.0  & 17.9 \pm 27.3 & 36.0 \\
		B & 20 &  NiftyReg & 105.0  & 12.8 \pm 11.6 & 135.0 \\
		\hline
		AC & 83 & FROG & $\bm{60.4} $ & $\bm{8.5\pm 6.9 } $& $\bm{1.3}$ \\
		AC & 83 & ANTs-S2 & 198.0  & 44.7 \pm 31.6 & 82.8 \\
		AC & 83 & NiftyReg & 64.0  & 9.3 \pm 9.3 & 324.0 \\
		\hline
		ABC &103 & FROG & 61.7  & ${8.7 \pm 7.0}$ & 1.9 \\
		
	\end{tabular}
\end{table}

\subsubsection{Overall}

Table \ref{tab:timings} shows timings for our algorithm as well as the average number of matches per volume for several scenarios.  Note that even if the number of keypoints per volume, is constant (20k), the average number of matches increases with the number of volumes. As an example, when registering group A, each keypoint is matched with $190k/20k=9.5$ points on average. But when registering groups A and C together, each keypoint is matched with $820k/20k = 41$ points on average.
In terms of memory footprint, our approach and NiftyReg never needed more than 10 GB, while we observed a peak memory consumption greater than 65 GB with ANTs at full resolution, and 34 GB when setting $sf=2$.

\begin{figure}
	\centering
	 Group A \hspace{2cm}   Groups A + C

	\subfigure[ANTS]{
		\includegraphics[width=0.3\linewidth]{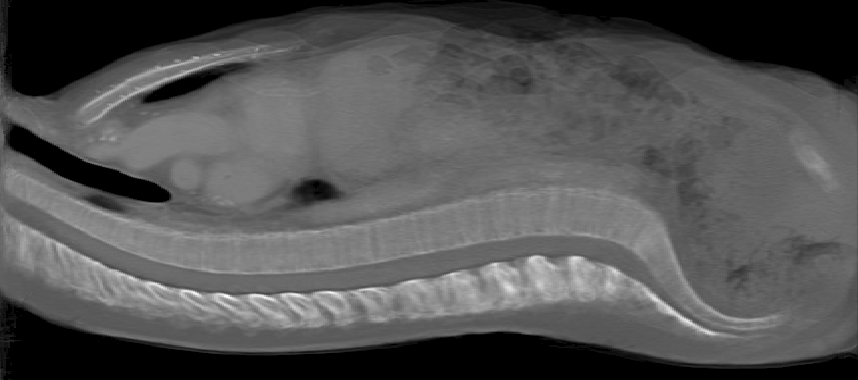}
		\includegraphics[width=0.3\linewidth, trim={0 0 0.2cm 0}, clip]{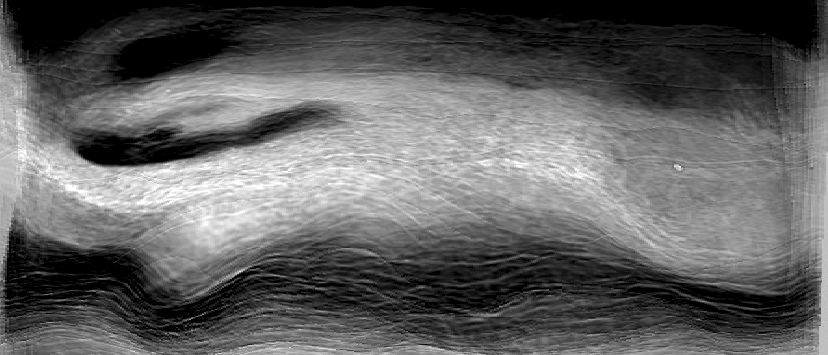}
	}

	\subfigure[NiftyReg]{
		\includegraphics[width=0.3\linewidth]{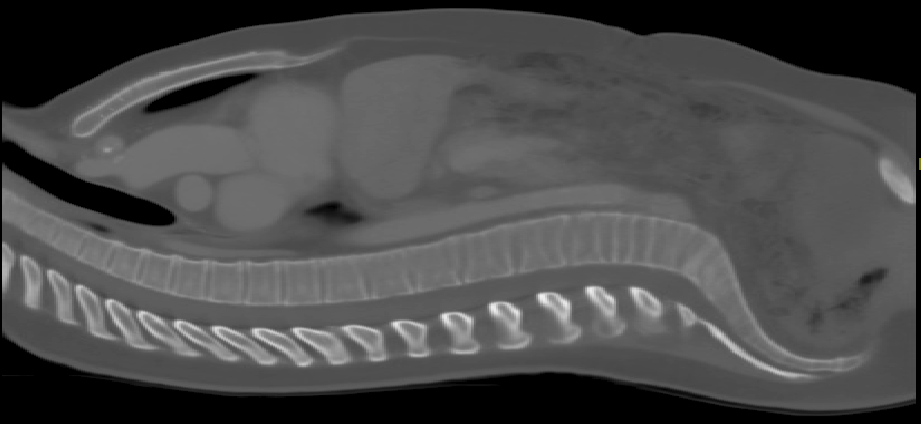}
	\includegraphics[width=0.3\linewidth, trim={0 0 0.8cm 0}, clip]{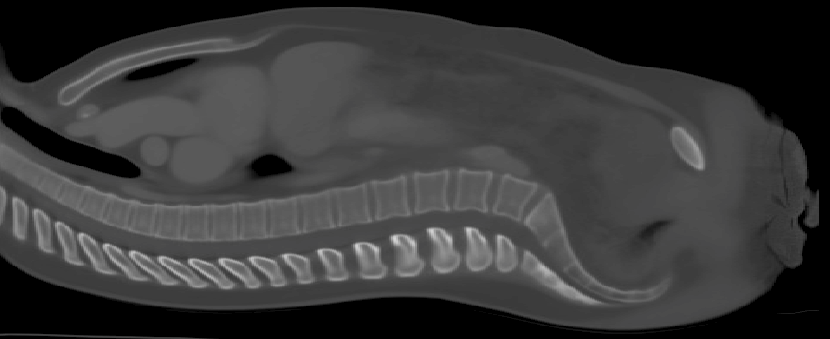}
}

	\subfigure[Ours]{
	\includegraphics[width=0.3\linewidth, trim={0.1cm 0 0.4cm 0}, clip]{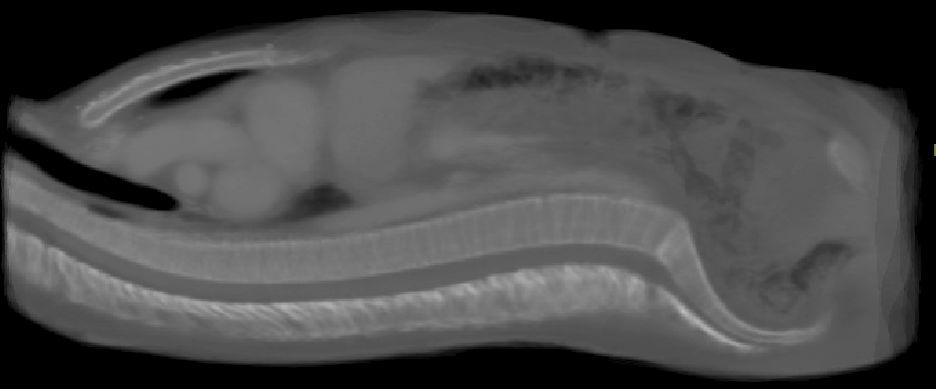}
	\includegraphics[width=0.3\linewidth]{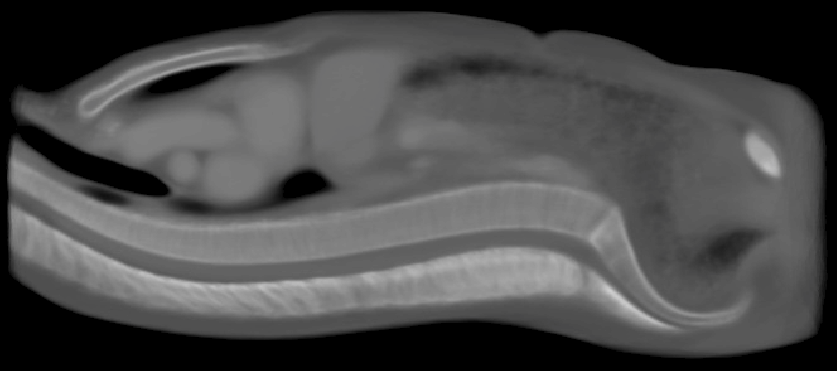}
}

\caption{Average images obtained with (a) ANTS, (b) NiftyReg, (c) Ours. Left : when registering group A. Right : when registering groups A+C.}
	\label{fig:average}
\end{figure}

\begin{figure}
\centering
\subfigure[NiftyReg ($10000100$)]{\includegraphics[width=0.49\linewidth, trim={1cm 2cm 2cm 5cm},clip]{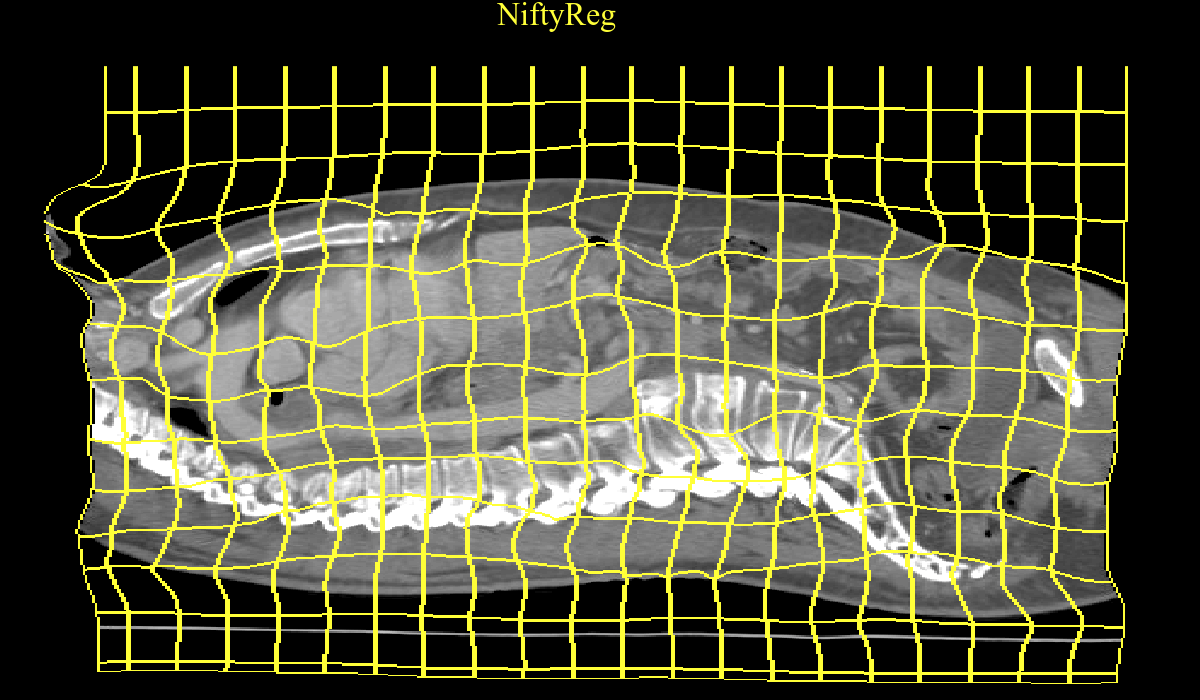}}
\subfigure[FROG ($10000100$)]{\includegraphics[width=0.49\linewidth, trim={1cm 2cm 2cm 5cm},clip]{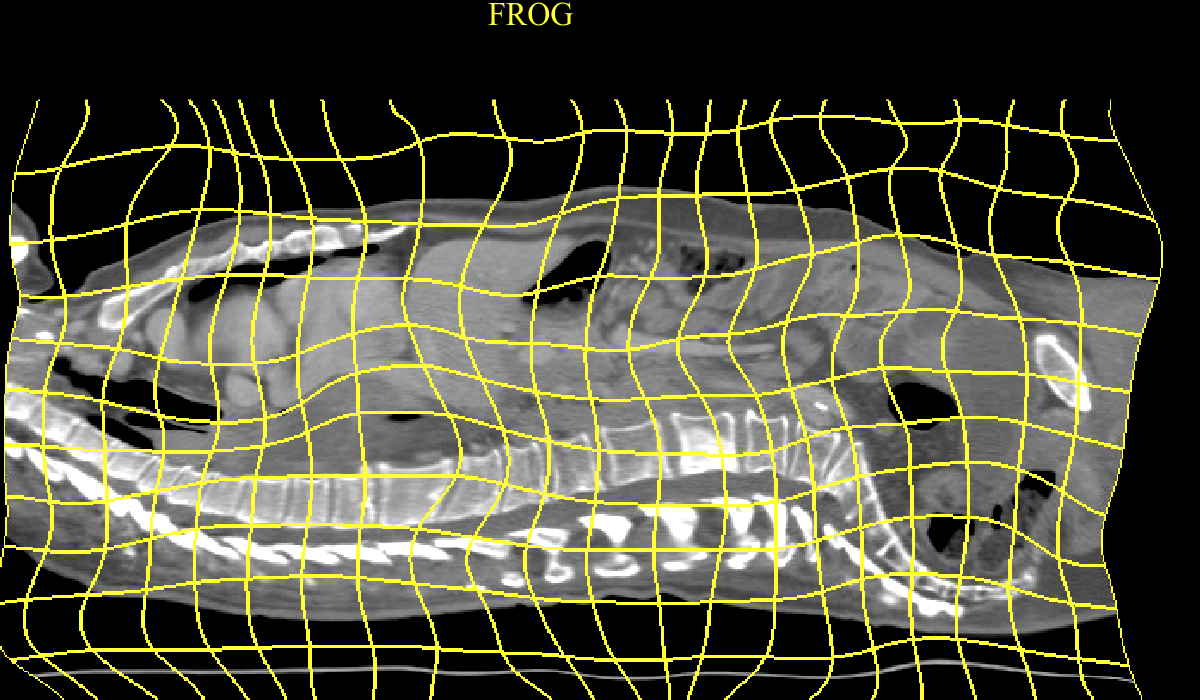}}
\subfigure[NiftyReg ($10000105$)]{\includegraphics[width=0.49\linewidth, trim={1cm 2cm 2cm 5cm},clip]{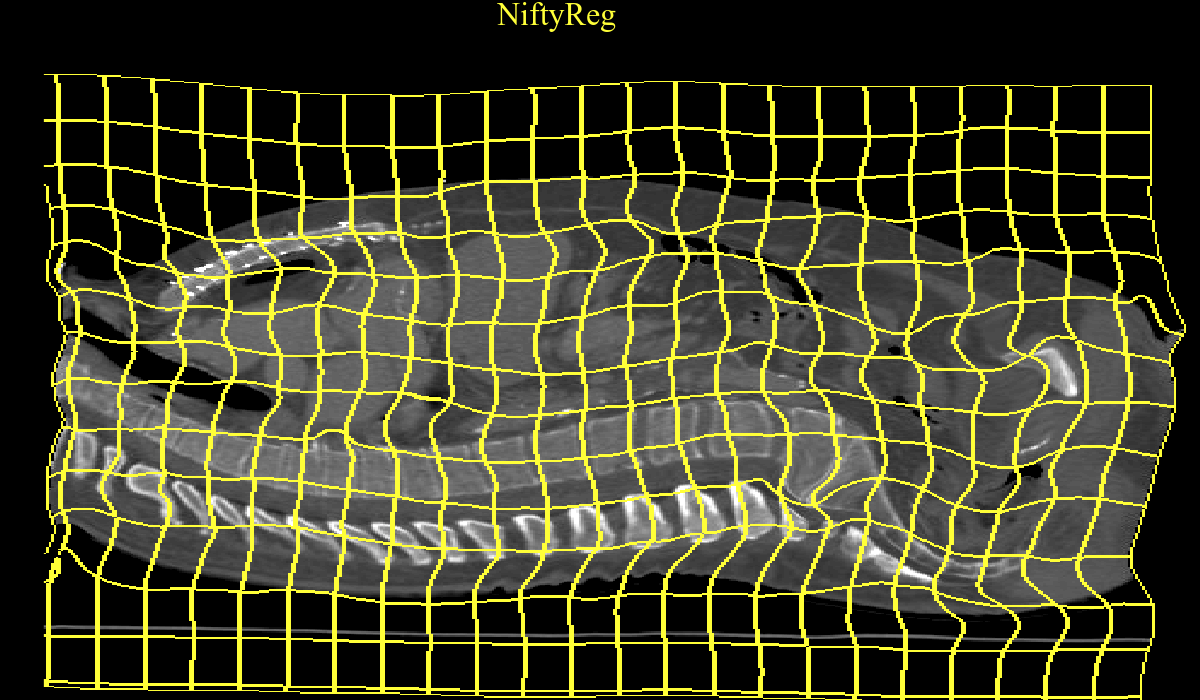}}
\subfigure[FROG ($10000105$)]{\includegraphics[width=0.49\linewidth, trim={1cm 2cm 2cm 5cm},clip]{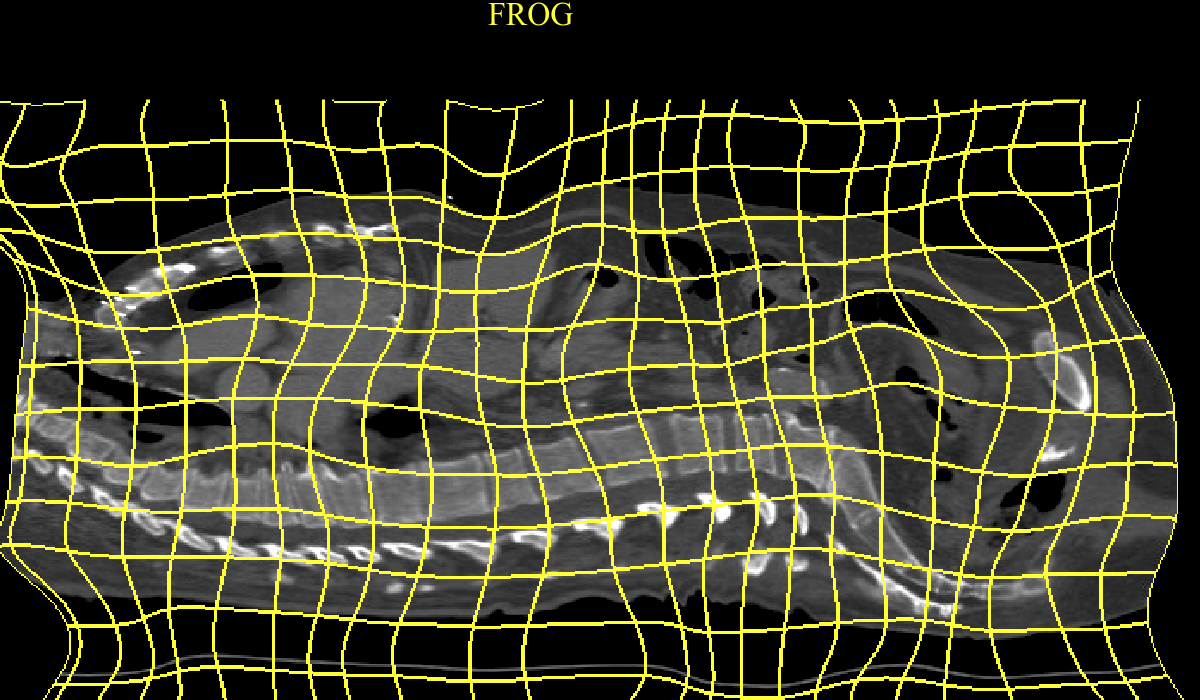}}
\caption{Comparison of transformed images for two individuals ($10000100$ and $10000105$ in VISCERAL Group A): Despite providing a sharper mean image (see figure \ref{fig:average}) NiftyReg sometimes yields transformations that are less realistic than FROG. This can be easily observed on some oversized lumbar vertebrae in $10000100$. FROG also sometimes yields unrealistic deformation, such as the chest boundary in $10000105$.}
\label{fig:image100}
\end{figure}

Finally, figure \ref{fig:average} shows the average of the registered images. From top to bottom: ANTs, NiftyReg and our algorithm, when registering Group A (left) and groups A and C (right). NiftyReg and ANTs provide a sharper image than our approach when registering group A. When registering groups A and C (83 images), the resulting average image is blurred for ANTs, whereas our approach remains robust to the number of images. NiftyReg still yields an average image sharper than ours. Hence, our approach is not very effective for template construction (average image). But, in contrast with NiftyReg and ANTs, our hubless approach does not need any average image during registration. Our average image is only a byproduct used for comparison, computed after the registration. Moreover, image-based criteria have been shown unreliable to evaluate registration accuracy \citep{rohlfing12}. We observed that even though NiftyReg yields a sharp mean image, individual transforms can be unrealistic. As an example, figure \ref{fig:image100} compares the images $1000100$ and $1000105$ from VISCERAL group A transformed by NiftyReg and FROG. The overlay grid reflects the deformation for all transforms. Some lumbar vertebrae are clearly oversized with NiftyReg on image $1000100$. On the other hand, FROG sometimes also deforms images in an unrealistic way, as shown on the chest boundary in image $1000105$. Finally, although our average image is not the sharpest, landmark distances (table \ref{tab:ours_vs_ants}) are on average lower with our approach.

We have also registered all three groups A, B, and C in less than 2 hours, with a reasonable error, which illustrates the ability of our approach to register more than 100 images, mixing images obtained with or without constrast agent.

\subsection{Using 3D-SIFT}

We have carried out an experiment with points defined with 3D-SIFT \citep{SIFT3D}, visible on line 7 in table \ref{tab:ours_vs_ants_global}. Keypoint extraction is much slower than our 3D-SURF approach, which results in a registration time of 4.2 hours, while yielding larger landmark distances (11.6mm for 3D-SIFT vs 9.0mm for 3D-SURF). This shows that our approach is able to exploit various keypoint types, and also illustrates the relevance of choosing 3D-SURF over 3D-SIFT, as reported initially by \cite{bay2006surf}.

\subsection{Limitation}

\begin{figure}[ht]
	\centering

	\includegraphics[height=0.3\linewidth]{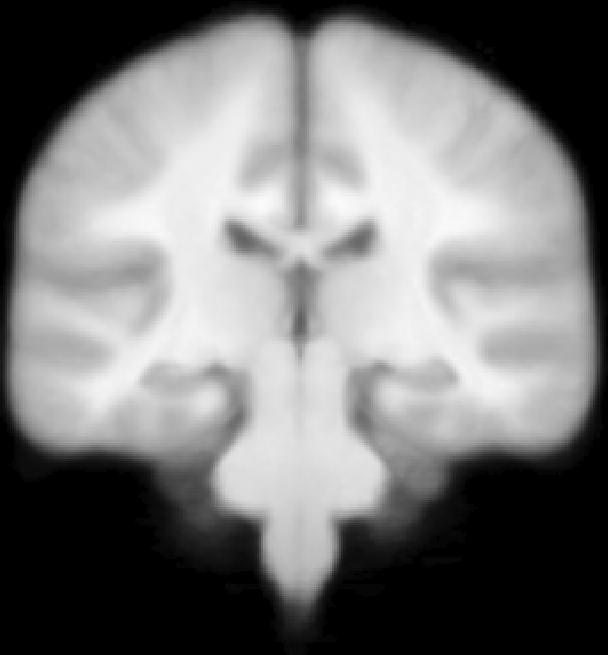}
	\includegraphics[height=0.3\linewidth]{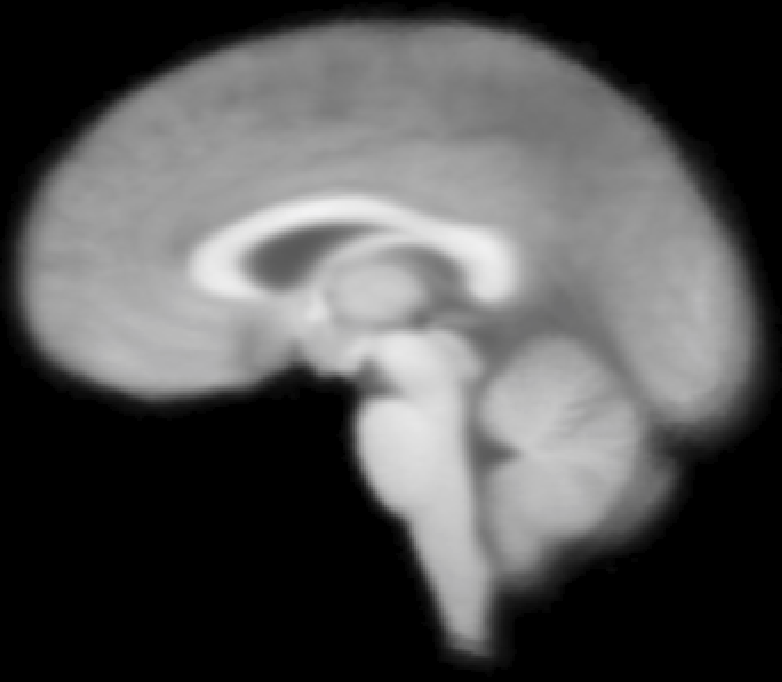}
	\vspace{12mm}
	\includegraphics[height=0.3\linewidth]{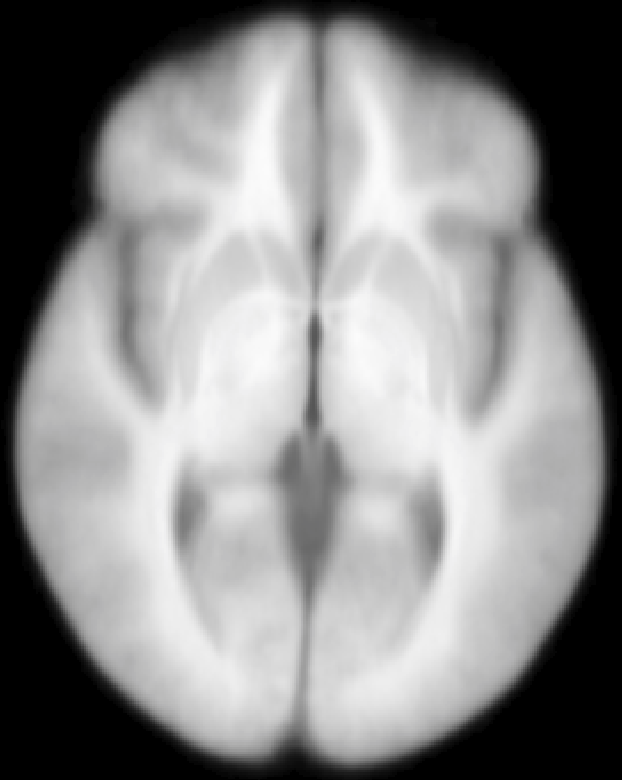}
	
	\caption{Average brain when registering the LPBA40 MRI brain dataset with our approach.}
	\label{fig:averageBrain}
\end{figure}

The limitation of our proposal is related to the requirement of a sufficient number of inlier keypoints in each cell of the grid at the highest resolution. In other words the size of the finest grid used is locally limited by the keypoint density. This results in a limit in registration accuracy. We observed this limitation in processing the MRI LONI Probabilistic Brain Atlas (LPBA40)  \citep{LPBA40}, containing 40 human brain images of resolution $256 \times 124 \times 256$. Processing time for the whole dataset is less than 10 minutes. The resulting average image is displayed in Figure \ref{fig:averageBrain}. The accuracy of our generic approach is not sufficient to correctly handle small structures inside the brain. Hence, our approach is not as accurate as algorithms dedicated to brain processing such as GLIRT\citep{wu2012feature}, which reports an average overlap ratio of $67\%$ (about $20\%$ higher than ours). Yet our algorithm converges in a short time, which illustrates the versatility of our approach. In practice, our approach is able to extract about 10k keypoints for each brain volume of LPBA40, while more that 100k points have been extracted from each volume of the VISCERAL database. Thus, for the VISCERAL dataset, we retained only the most significant points, discarding weak points. This was not possible for the LPBA40 database. This comes from the fact that brain MRI images are less sharp and contain fewer voxels than full boddy CT scans. Hence, a possible improvement could be to develop a keypoint extractor more adapted to brain MRI, which could yield much more keypoints and therefore increase registration accuracy.

\begin{table}
	\small
	\caption{Number of matches per volume and processing times for our algorithm. $n$ is the number of registered volumes, $|\mathcal{M}|/n$ the number of matches per volume. Columns I, II and III indicate the time spent (in minutes) for keypoint extraction, matching and optimization, respectively.}
	\label{tab:timings}
	\centering
	\begin{tabular}{|r|r|r|r|r|r|r|}
		\bf Data &\bf  $n$  &\bf  $|\mathcal{M}|/n$ &\bf  I (mn) &\bf  II (mn) &\bf  III (mn) &\bf  Total (mn)\\
		\hline
		A    & 20  &     190k        & 6.3 & 1.5 & 2.2 & 10.0 \\
		B   & 20  &	190k	     & 24.4 & 1.5 & 2.6 & 28.5 \\
		AC   & 83  &	820k	     & 30.2 & 28.5 & 20.6 &  79.3\\
		ABC  & 103 &      1019k       & 52.8 & 31.3 & 32.8 & 116.9 \\
	\end{tabular}
\end{table}

\section{Conclusion}
\label{sec:discussion}

Our algorithm is able to register CT and MRI volumes in reasonable time and low memory consumption.
Experimental results illustrate its robustness to variability, and comparisons with NiftyReg and ANTs show that our algorithm yields lower mean landmark distances within a much shorter computation time. Our approach is currently suited for large databases of relatively high resolution images. As an example, brain databases generally contain images with $256 \times 256 \times 256$ voxels, but our approach is not as accurate as state-of-the-art voxel-based approaches for these resolutions. A possible improvement could be a keypoint extractor dedicated to MRI images with fewer voxels and fewer sharp features. But many applications could already benefit from our approach. As an example, the low complexity and memory requirement of our algorithm allows to use a full groupwise nonrigid registration algorithm as a preliminary task for applications such as atlas based segmentation, machine learning classification and longitudinal studies. A preliminary application using FROG is proposed in \cite{franchi2019}, where we predict patient gender from 3D CT images using groupwise registration.

As depicted in figure \ref{fig:hubless}, we use a full graph approach. As a consequence, the number of matches grows faster than the number of images. Hence, the number of matches per images increases with the number of images (see Table \ref{tab:timings}). While adding more images has the potential to produce more matching keypoints to improve registration accuracy, our results did not demonstrate statistically significant improvement.

Our method is oblivious to point cloud sources and is not restricted to SURF keypoints and splines. Hence, investigating keypoint descriptors tailored to medical images and richer transform models are two very interesting research perspectives for our work. Computing features with machine learning seems a promising option in this context.

As our algorithm provides a robust approach able to handle keypoints between different patients without needing dense voxel information during optimization, it paves the way towards fast algorithms able to deal with large data sets. 
We showed that although our approach exhibits quadratic complexity and memory footprint, it is able to register more than one hundred images in a short time (less than 2 hours). With our current hardware (128GB RAM), we project to be able to register as many as 1000 volumes with FROG, in about 125 hours. Note that this would still be less than the time taken by NiftyReg to register 83 volumes.  Scaling the algorithm to even larger data sets (more than 10000 patients) is still a challenging problem.

\section*{Acknowledgment}
Many thanks to the VISCERAL Consortium \citep{VISCERAL} (www.visceral.eu) for allowing us to compute evaluations using the VISCERAL data set.


\bibliography{biblio}

\end{document}